\documentclass[preprint,11pt]{elsarticle}
\usepackage{tikz}
\usepackage{pgfplots}
\pgfplotsset{compat=1.18}
\usepackage{pgfplotstable}
 \usepackage{longtable}
\usepackage{booktabs}

\usepackage{adjustbox}
\usetikzlibrary{shapes, arrows, positioning}
\usepackage{url}
\usepackage{pifont}
\usepackage{graphicx}
\usepackage{setspace}
\usepackage{amsmath}
\usepackage{amsthm}
\usepackage{amssymb}
\usepackage{hyperref}
\usepackage{algorithm}
\usepackage{algpseudocode}
\usepackage{algorithmicx}
\usepackage{caption} % <-- Add this
\usepackage{float}
\usepackage[utf8]{inputenc}
\usepackage{tcolorbox}
\usepackage{tikz}
\usepackage{xcolor}
\usepackage{booktabs}
\usepackage{listings}

\definecolor{codegray}{gray}{0.95}

\lstdefinestyle{gurobicode}{
    backgroundcolor=\color{codegray},
    basicstyle=\ttfamily\small,
    keywordstyle=\color{blue!50!black}\bfseries,
    commentstyle=\color{green!40!black},
    stringstyle=\color{orange!60!black},
    numbers=left,
    numberstyle=\tiny,
    numbersep=6pt,
    breaklines=true,
    breakatwhitespace=true,
    tabsize=2,
    frame=single,
    framesep=3pt,
    showstringspaces=false,
    keepspaces=true,
    columns=fullflexible,
}

\usepackage[margin=1in]{geometry}
\tcbset{
  promptbox/.style={
    colback=black!5!white,
    colframe=black!75!black,
    fonttitle=\bfseries,
    coltitle=white,
    boxrule=0.5mm,
    arc=2mm,
    top=2mm,
    bottom=2mm,
    left=2mm,
    right=2mm,
    title=Prompt
  }
}
\tcbset{promptboxenv/.style={
  breakable,
  before skip=8pt,
  after skip=8pt,
  pad at break*=1mm
}}

\lstdefinestyle{mypython}{
    language=Python,
    basicstyle=\ttfamily\small,
    numbers=none,
    backgroundcolor=\color{gray!10},
    frame=single,
    mathescape=true,
    morekeywords={self},
    commentstyle=\color{green},
    keywordstyle=\color{blue},
    stringstyle=\color{red}
}


\begin{document}

\theoremstyle{plain}
\newtheorem{theorem}{Theorem}[section]
\newtheorem{proposition}[theorem]{Proposition}
\newtheorem{lemma}[theorem]{Lemma}
\newtheorem{corollary}[theorem]{Corollary}

\theoremstyle{definition}
\newtheorem{definition}[theorem]{Definition}
\newtheorem{assumption}[theorem]{Assumption}

\newtcolorbox[auto counter, number within=section]{promptboxenv}[2][]{%
    promptbox,             % your style
    title={#2},
    label={#1},
}

\singlespacing
\begin{frontmatter}

\title{Large Language Model-Based Automatic Formulation for Stochastic Optimization Models}

\author[1]{Amirreza Talebi\corref{cor1}}
\ead{talebi.14@osu.edu}
% \author[1]{G\"uzin Bayraksan}

% \ead{bayraksan.1@osu.edu}

\cortext[cor1]{Corresponding author}
% \fntext[fn1]{Email: bayraksan.1@osu.edu}

\affiliation[1]{{Department of Integrated Systems Engineering,},
            addressline={The Ohio State University}, 
            city={Columbus},
            postcode={43210}, 
            state={OH},
            country={USA}}

%% Abstract
\begin{abstract}

This paper presents an integrated systematic study of the performance of large language models (LLMs), specifically ChatGPT, for automatically formulating and solving Stochastic Optimization (SO) problems from natural language descriptions. Focusing on three key categories, individual chance-constrained models, joint chance-constrained models, and two-stage stochastic mixed-integer linear programming models, we design several prompts that guide ChatGPT through structured tasks using chain-of-thought and agentic reasoning.
We introduce a novel soft-scoring metric that evaluates the structural quality and partial correctness of generated models, addressing the limitations of canonical and execution-based accuracy metrics. Across a diverse set of SO problems, GPT-4-Turbo achieves better partial scores than GPT-3.5 variants except for individual chance-constrained problems. Structured prompts significantly outperform simple prompting, reducing extra-element generation and improving objective matching, although extra-element generation remains a nontrivial task.
Our findings reveal that with well-engineered prompts and multi-agent collaboration, LLMs can facilitate SO formulations, paving the way for intelligent, language-driven modeling pipelines for SO in practice.

\end{abstract}

\begin{keyword}
stochastic optimization, ChatGPT, chain-of-thought,  multi-agent prompting, soft scoring
\end{keyword}

\end{frontmatter}

\section{Introduction}

\subsection{Motivation}
At the intersection of decision science and artificial intelligence in a rapidly changing research landscape, large language models (LLMs) such as ChatGPT \cite{chatgpt2025} offer new opportunities to automate operations research (OR) tasks. Prior research has shown promise in translating natural language (NL) descriptions of mathematical problems into formulations, especially for practitioners (e.g., \cite{shojaee2025llm, tsoukalas2024putnambench, ying2024lean}). In the optimization domain, most of this work has concentrated on deterministic optimization formulations or robust optimization (e.g., \cite{bertsimas2024robust,linzi2025systems,xiao2023chain}).

In contrast, Stochastic Optimization (SO), particularly in problems involving chance constraints, and two-stage stochastic mixed-integer linear programming models (SMILP-2), has received less attention, despite its central role in modeling uncertainty in real-world decision-making problems.

Many real-world decision problems involve uncertainty, which necessitate the use of complex SO models.
For example, SMILP-2 formulations govern long-term commitments followed by adaptive recourse actions, as in supply-chain design \cite{reis2023two}, inventory management \cite{cavagnini2024two}, or disaster logistics \cite{rodriguez2023two}. 
Chance-constrained variants capture risk-controlled feasibility with broad applications in real-world engineering problems such as portfolio optimization \cite{tagawa2024prospect}, air traffic flow management \cite{pang2024chance}, vehicle routing \cite{dinh2018exact}, and renewable energy integration into power systems \cite{guo2021chance, tian2024chance}. More real-world applications of SO, such as hydraulic control of underground water plume contamination, can be found in \cite{wallace2005applications}. 
Table~\ref{tab:stochastic_examples} summarizes representative applications spanning these categories.

While deterministic optimization models can often be specified directly from known parameters, SO introduces additional layers of uncertainty, recourse structures, and probabilistic feasibility in reality that make these models more challenging to formulate and implement in practice.
Despite their prevalence, constructing SO formulations requires knowledge in multiple fields, such as optimization and probability, which limits their further use and applicability for many practitioners. 

LLMs could help remove such barriers for practitioners. However, this needs to be done carefully and systematically to ensure correctness. 
This work is motivated by making SO more accessible to practitioners by enabling LLMs to automatically formulate and evaluate these classes of stochastic models from NL descriptions, bridging NL reasoning and formal optimization.
Leveraging LLMs for modeling complex SO problems in a structured manner is crucial, as they provide quick, step-by-step guidance and clear explanations, making complex modeling and implementation tasks more comprehensible. This capability enables rapid testing of multiple scenarios and real-time analysis, while enhancing decision-making with improved explainability and iterative refinement. However, even doing this in practice is not straightforward. Correct and structured prompts are needed, and subsequently, there is a need to evaluate the performance of the prompts and state-of-the-art LLMs, particularly ChatGPT, and their accuracy in formulating and coding SO problems.   

\begin{table}[H]
\tiny 
\centering
\caption{Representative real-world applications of SO by model type.}
\begin{tabular}{p{3cm} p{2.2cm} p{2cm} p{5.8cm}}
\toprule
\textbf{Real world problem} & \textbf{Model type} & \textbf{Reference} & \textbf{Relevance to SO} \\
\midrule

\textbf{Inventory management} &
SMILP-2 &
\cite{cavagnini2024two} &
First stage: decisions regarding the target level of inventory in the bike sharing system, second stage: re-balancing the inventory levels to meet uncertain demand\\

\textbf{Supply chain network design} &
SMILP-2 &
\cite{reis2023two} &
First stage: contract decisions, such as the amount of grains purchased; second stage: resource flows after demand, price, and revenue under uncertainty realization. \\

\textbf{Disaster relief / Humanitarian logistics} &
SMILP-2 &
\cite{rodriguez2023two} &
First stage: selection of suppliers, agencies, critical facilities, and pre-positioning of supplies; second stage: resource allocation, relief distribution, and supply of extra resources. \\

\textbf{Hydraulic control of underground water plume contamination}& SMILP-2& \cite{wallace2005applications} & First stage: where to build a well; second stage: handling hydraulic conductivity, plume containment constraints, and pumping rates.\\

\textbf{Air traffic flow management} &
 Chance-constrained &
\cite{pang2024chance}&
Control the probability of loss of separation at shared way-points (avoiding conflicts between airplanes at way-points with high probability) \\

\textbf{Energy unit commitment with renewables} &
Chance-constrained &
\cite{guo2021chance} &
Wind power load balance constraints must jointly hold with high probability at all periods for each bus.\\

\textbf{Integrating renewable energy into power grids} &
Chance-constrained &
\cite{tian2024chance}&
Enforces that all spinning reserve constraints hold simultaneously with a specified probability threshold. \\

\textbf{Chance-constrained vehicle routing problem} &
Chance-constrained &
\cite{dinh2018exact} &
Build minimum length routes so that, based on customer demand and the vehicle capacity, such that every customer is met exactly once, and each vehicle's capacity is exceeded with a specified probability threshold. \\

\textbf{Portfolio optimization} &
Chance-constrained &
\cite{tagawa2024prospect} &
Maximize profit given that the return at the end of a period is less than a value with a specified probability threshold. \\

\bottomrule
\end{tabular}
\label{tab:stochastic_examples}
\end{table}

\subsection{Research gap and novelty}

In addition to the gap in considering automatic formulation of SO problems mentioned earlier, existing prompt-based and multi-agent frameworks (e.g., \cite{linzi2025systems, shojaee2025llm, tsoukalas2024putnambench}) evaluate canonical\footnote{Canonical accuracy measures whether the generated formulation matches a canonical representation of the optimization problem (a standardized linear programming (LP) model form, e.g., see \cite[Figure 2]{Ramamonjison2023NL4OptCF}).} accuracy quality but do not assess to what extend LLMs can formulate SO problems containing probabilistic constraints, and recourse actions.

This absence of benchmarks, evaluation metrics, and structured methodologies for SO formulation constitutes a key gap in the current literature. Therefore, there is currently no systematic methodology for evaluating the correctness of LLM-generated SO formulations.

This paper fills this gap by introducing a new framework that combines prompt engineering, multi-agent reasoning, and algebraic and structural evaluation for formulating two commonly used SO models: chance-constrained optimization models and SMILP-2.  

We highlight several novel outcomes:

\begin{enumerate}[(i)]
    \item the discovery of consistent reasoning patterns across GPT-4 and three models when handling expectation-based objectives and probabilistic constraints;
    \item the design of a multi-agent prompting system that relatively mitigates hallucinations by decomposing model synthesis into specialized roles, and multiple reviewers, and
    \item the development of a soft scoring metric that provides partial scores for correct components of LLM-generated code outputs.  
\end{enumerate}

\subsection{Contributions}

This paper makes the following key contributions:
\begin{enumerate}[(i)]
    \item First systematic application of LLMs to formulate SO problems, including SMILP-2 and chance-constrained formulations, directly from NL descriptions, helping practitioners. 
    \item Customized prompting strategies tailored for practitioners, including stepwise reasoning and problem-specific instructions.
    \item A soft scoring metric for model evaluation that accounts for structural similarity, notational variation, and component-level permutations, offering a more robust alternative to canonical or execution-based scoring.
    \item Building several test instances, including NL descriptions of problems, input data, ground truth code, and output optimal objective solution.
    \item Extensive experimental evaluation across GPT-3.5 and GPT-4 variants and several prompt designs, analyzing performance trends, error types, and consistency and reproducibility across problem categories.
\end{enumerate}

Our experiments demonstrate that GPT-4 models, particularly GPT-4-Turbo, achieve better structural matching scores than GPT-3 models. Additionally, structured prompting strategies consistently outperform simple prompts across various problem classes. The proposed scoring metric further captures partial correctness that canonical and execution-based metrics miss, revealing meaningful progress even when exact accuracy remains zero.

\section{Related work}\label{sec:lit}

\subsection{Paradigms for automatic formulation of OR problems using LLMs}

Optimization modeling and SO have traditionally required significant domain expertise, but LLMs now offer a path toward automation, although it is in an early stage. Two main paradigms exist leveraging LLMs to formulate OR problems: (i) prompt-based methods, which guide pretrained LLMs via structured prompts or multi-agent workflows, and (ii) learning-based methods, which fine-tune models\footnote{Fine-tuning updates parameters of a pretrained model to make the model focus on a specific domain. Some variants, such as LoRa \cite{hu2021loralowrankadaptationlarge}, only fine-tune a subset of parameters to save time and costs.} on synthetic datasets (e.g., \cite{jiang2024llmopt}) to generate variations of mathematical formulations, problem instances, input data (data synthesis), and so on \cite{zhang2025systematic}. 
These two paradigms, in particular the prompt-based approach, build the grounds for the following line of research in LLM-based formulation of OR problems.

\subsection{Prompt-based formulation of OR problems}

Prompt-based approaches use NL problem descriptions to generate or refine optimization formulations, heuristics, or solver code on pre-trained LLMs. Zero-shot prompting relies solely on pretrained knowledge, i.e., a single command to render the results given a problem, while few-shot prompting\footnote{One-shot prompt is a variant of few-shot prompts with only one example \cite{brown2020language}.} provides examples that implicitly define structure and notation \cite{brown2020language}. For mathematically intensive tasks,  Chain-of-Thoughts (CoT) prompting elicits intermediate and step-by-step reasoning to reduce shortcut errors and clarify logic \cite{wei2022chain}. The Chain-of-Experts (CoE) framework, on the other hand, decomposes problems into specialized roles such as extractor, formulator, and verifier, coordinated by a central controller \cite{xiao2023chain}.  Extending this idea, multi-agent prompting lets several LLM agents critique, repair, and validate each other’s work or invoke external tools for more reliable formulations \cite{li2023camelcommunicativeagentsmind}. 

In this paper, we developed zero-shot and few-shot CoT prompts for a range of SO problems. We have also included prompts with instructions inspired by the work of \cite{bertsimas2024robust} in SO, which also improves the performance of zero- and few-shot prompts \cite{liu2023visual}. There are other prompting methods, such as Reflection, progressive hinting, Tree-of-Thought, ReAct, etc. We refer the reader to \cite{xiao2023chain} for formal definitions of these prompts. While some of these methods outperformed CoT on deterministic mixed-integer linear programming (MILP) problems, they were capable of solving only a few simple deterministic optimization problems. CoT provides the most controlled and interpretable baseline for evaluating whether LLMs can formulate SO models as applied in the recent literature (e.g., \cite{bertsimas2024robust}), independent of reasoning heuristics or external tool usage.

\subsection{Deterministic (MI)LP problem formulations using LLMs}

As for examples of formulating deterministic (MI)LP problems, initially,   \cite{ramamonjison2022augmenting} proposed OptGen, an augmented intelligence system that automatically generates LP models from NL problem descriptions. OptGen uses a two-stage pipeline, mapping text to an intermediate representation and then converting it into a canonical LP, and introduced the first dataset of LP Word Problems (LPWP). Building on this, \cite{Ramamonjison2023NL4OptCF} organized the NL4Opt Competition, where they also proposed a canonical-form metric to assess deterministic LP problem formulations generated by LLMs. Subsequently, \cite{Li2023SynthesizingML}  extended the NL4Opt dataset and designed a 3-stage MILP synthesis pipeline (variables, constraints, and model) for the formulation of deterministic MILP problems. 
\cite{ahmed2024lm4opt} evaluated LLMs on the NL4Opt dataset. Similar to this paper, they found that few-shot prompting outperforms zero-shot prompting, and fine-tuning using instructions can reduce hallucinations of LLMs, but the size of instructions is important. \cite{ghiani2025improving} integrated LLMs with optimization algorithms to create adaptive decision-support systems. Their method interprets NL instructions, adjusts algorithmic components, and explains decisions. In contrast, we have developed a scoring metric evaluating the performance of LLMs on SO problems rather than decision-making criteria. 
In this work, for the first time, we developed and gathered SO problems for testing LLMs, whereas the mentioned literature is designed for deterministic optimization problems.

\subsection{Multi-agent and heuristic OR frameworks with LLMs}

Several works extended the formulation of deterministic (MI)LP problems with single-agent prompts to multi-agent prompting.

The reference \cite{mostajabdaveh2024optimization} introduced a multi-agent framework with relation-identification and verification agents that operate without solver execution on deterministic LP, MILP, and quadratic programming (QP) problems. Their evaluation, however, depends on embedding-based similarity between ground-truth and generated codes and comments, i.e., converting comments and codes into vectors and finding similarity of these embeddings by the cosine-similarity approach. 
The OptiMUS framework by \cite{ahmaditeshnizi2024optimus} employs a multi-agent architecture to formulate and solve deterministic (MI)LP problems from NL descriptions. Similar to this paper, they introduced a single evaluator agent to capture and report errors of generated code by LLMs. \cite{ahmaditeshnizi2025optimus03usinglargelanguage} extended OptiMUS. In particular, they included a self-reflective error correction method, enhancing the LLM's output for deterministic (MI)LP problems formulation.  \cite{zhang2025or} presented OR-LLMAgent, a reasoning-based framework that decomposes OR workflows into modeling, code generation, and debugging agents. On LP problem dataset benchmarks, they achieved over 7\% accuracy improvement, confirming the value of task decomposition or a multi-agent framework.
This literature, unlike this paper, does not contain multiple reviewers, stochastic problems, and granular evaluation metrics or instructions-included prompts.

As for designing heuristics using LLMs, 
\cite{liu2024large} proposed an LLM-driven Evolutionary Algorithm (LMEA) which constructs prompts to guide LLM for choosing algorithmic decisions, such as parent solutions in a genetic algorithm, or \cite{li2025llm} addressed the lot-streaming hybrid job shop scheduling problem (LHJSV) using a memetic algorithm guided by an LLM-designed heuristic, combining problem decomposition with prompt learning. \cite{10965770} proposed an LLM-based framework that autonomously designs multi-objective evolutionary algorithms (MOEAs). The framework uses runtime and execution errors to refine prompts and reproduction methods automatically by LLMs. Although the heuristic approach in these papers inherently contains stochasticity, they do not address the SO problem formulations. 

\subsection{SO and decision-making under uncertainty using LLMs}

A limited literature has worked on formulating SO problems by LLMs. For example,
\cite{bertsimas2024robust} showed that ChatGPT can formulate and solve adaptive robust optimization (ARO) problems under uncertainty using CoT prompts in which instructions are provided. Their single-agent design does not address stochastic frameworks such as SMILP-2 or chance-constrained problems. \cite{kumar2025performance} evaluated LLMs’ capabilities in stochastic modeling using graduate-level exam problems and the SimOpt benchmark, showing near-expert proficiency in decision-making under uncertainty. In particular, they evaluated student-submitted solutions and ground-truth using LLMs. The work of \cite{liu2024dellma} developed a Decision-making Large Language model Assistant (DeLLMa). In this framework, given a problem context, a user goal, and a set of actions, the LLM agent is asked to develop a policy function, a probabilistic function mapping states to actions. Furthermore, the agent is asked to develop a utility function assigning values to states and actions. Finally, it will return the action maximizing the utility function.
Another work focusing on stochastic processes rather than SO problem formulations, \cite{rasal2024optimal}, introduced an LLM framework that embeds a Monte Carlo simulation module to run simulations on an optimization problem obtained by LLMs for a problem description. Then, an embedded optimization engine analyses the simulation outputs and renders an optimal solution. 

In contrast to these approaches, which focus on robust optimization, decision-making under uncertainty, or simulation-based solution methods, this paper aims at automatic formulation of SMILP-2 and chance-constrained SO models directly from NL descriptions using LLMs.

\subsection{Evaluation metrics and reliability of LLMs for OR problems}

A separate line of work uses LLMs for the reliability and interpretability of the outputs. \cite{chen2024diagnosing} developed OptiChat, a GPT-4-based framework for diagnosing infeasible MILP models by explaining structure, locating errors, and proposing corrective actions. 
Another aspect of our work is the granular evaluation metric. Evaluation methods in the literature largely depend on canonical and execution accuracy \cite{ prasath2023synthesismathematicalprogramsnatural, ramamonjison2022augmenting}. Canonical accuracy counts matched elements between generated and ground-truth models but fails to handle permutation and notation variations. Execution accuracy, while practical, cannot capture structural correctness when different formulations yield identical optimal solutions. \cite{linzi2025systems} addressed permutation issues via bipartite-graph matching, but their method remains sensitive to notation inconsistencies, computational overhead, and partial-correctness ignorance. These limitations motivate the need for granular evaluation metrics for SO formulations using LLM.

Table~\ref{tab:literature_review} provides a summary of related works. Existing frameworks either remain fully deterministic optimization or rely on ARO or stochastic heuristic algorithms, such as the design of genetic or evolutionary methods using LLMs. They lack granular evaluation metrics or multi-agent reviewer frameworks. Besides, we consider a variety of prompting methods for SO problems that integrate instructions and are easy to implement and adapt for practitioners.

{\scriptsize 
\begin{longtable}{p{3.2cm} p{5.2cm} p{7.2cm}}
\caption{Summary of related literature on LLM-based optimization and formulation.}
\label{tab:literature_review}\\
\toprule
\textbf{Category \& reference} & \textbf{Main contribution} & \textbf{Notes} \\
\midrule
\endfirsthead

\toprule
\textbf{Category \& reference} & \textbf{Main contribution} & \textbf{Notes} \\
\midrule
\endhead

\midrule
\multicolumn{3}{r}{\emph{Continued on next page}} \\
\endfoot

\bottomrule
\endlastfoot

\multicolumn{3}{l}{\textbf{NL-to-optimization formulation frameworks (deterministic (MI)LP problems)}} \\

\cite{ramamonjison2022augmenting} & Introduced OptGen, a two-stage pipeline mapping NL to canonical LPs; created LPWP dataset. & Focused on deterministic LP problems. \\

\cite{Ramamonjison2023NL4OptCF} & Organized NL4Opt competition defining entity recognition and logical-form generation tasks with a canonical accuracy metric. & Focused on deterministic LP problems. \\

\cite{ghiani2025improving} & Integrated LLMs with optimization algorithms for adaptive decision-support and interpreting NL instructions, adjusting algorithmic
components, and explaining decisions. & Focused on deterministic (MI)LP, and heuristics containing stochasticity.\\

\cite{Li2023SynthesizingML} & Designed a 3-stage MILP synthesis pipeline (variables, constraints, and model). & Focused on deterministic (MI)LP problems.\\

\cite{ahmed2024lm4opt} & Benchmarked LLMs for NL2OPT translation; found GPT-4 best in one-shot settings. & Focused on deterministic (MI)LP problems. \\

\multicolumn{3}{l}{\textbf{Multi-agent frameworks}} \\

\cite{mostajabdaveh2024optimization} & Proposed multi-agent relation-identification and verification agents,  without solver execution; introduced new dataset. & Relies on embedding-based similarity rather than algebraic model checking. Focused on deterministic (MI)LP, and QP problems. \\

\cite{ahmaditeshnizi2024optimus} & Developed OptiMUS, a modular multi-agent MILP formulation and solving framework using Python/Gurobi. & Focused on deterministic MILPs. \\

\cite{ahmaditeshnizi2025optimus03usinglargelanguage} & Extended OptiMUS with debugging and refinement for complex NL MILP problems. & Focused on deterministic (MI)LP problems. \\

\cite{zhang2025or} & Introduced OR-LLMAgent-reasoning-based for modeling, code generation, and debugging; developed BWOR dataset. & Focused on reasoning and deterministic (MI)LP problems. \\

\cite{xiao2023chain} & Proposed CoEs architecture with feedback and reflection for (MI)LP problems. & Focused on deterministic (MI)LP problems based on canonical evaluation. \\

\multicolumn{3}{l}{\textbf{LLM-assisted heuristic, evolutionary}} \\

\cite{10965770} & Introduced LLM-driven MOEA design using error-driven dialogue, prompt-based evolution, and dynamic strategy selection for tuning heuristics. & Focused on evolutionary heuristic algorithms containing stochasticity. \\

\cite{li2025llm} & Applied LLM-designed heuristics to solve LHJSV scheduling via memetic algorithms. & Covers heuristics containing stochasticity and deterministic problems.\\

\cite{liu2024large} &  Proposed LMEA constructing prompts to guide LLM for choosing algorithmic decisions, such as parent solutions in a genetic algorithm. & Focused on LLM hyperparameter tuning and heuristics containing stochasticity. \\

\multicolumn{3}{l}{\textbf{Stochastic and simulation-oriented}} \\

\cite{bertsimas2024robust} & Demonstrated ChatGPT solving ARO with CoT prompts and instructions. & Single-agent; Focused on ARO. \\

\cite{kumar2025performance} & Evaluated LLMs in stochastic modeling and simulation-optimization using SimOpt. & Assessment-focused rather than an SO modeling framework. \\

\cite{liu2024dellma} & Introduced DeLLMa, integrating utility theory for decision-making under uncertainty. & Focused on reasoning-level rather than SO problem formulation.\\

\cite{rasal2024optimal} & Introduced an LLM framework that embeds a Monte Carlo simulation module to run simulations on an optimization problem obtained by LLMs for a problem description.  &  Focused on Monte Carlo simulation for generated formulations using LLMs, assessing results based on an embedded optimization engine.\\

\multicolumn{3}{l}{\textbf{Evaluation metrics, debugging, and verification}} \\

\cite{chen2024diagnosing} & Developed OptiChat, a GPT-4 system diagnosing infeasible optimization models. & Focused on post-solution debugging and deterministic (MI)LP problems. \\

\cite{linzi2025systems} & Used bipartite-graph matching for canonical-to-ground-truth model comparison. & Focused on evaluation metric based on bipartite graph matching for generated and ground truth models. \\

\cite{madaan2023self} & Introduced feedback-based refinement and self-correction for LLM reasoning tasks. & Applied to general reasoning in LLMs rather than testing in the OR domain. \\

\end{longtable}}

\section{Approach}\label{sec:approach}
 In this section, we introduce formulations, prompts, justification for our framework design, motivating examples, and formal representation and justification of the scoring system. 
 For each problem category, a general formulation is presented below in its standard analytical form, which is used as a reference and as an input structure for prompts with instructions. These materials have been adapted from \cite{birge1997introduction, shapiro2021lectures}. It is to be mentioned that our test cases include cases in which integer variables also exist. 

Throughout this paper, the $i^{th}$ row of a matrix $A$ is denoted by $A_{i\cdot}$, and the $i^{th}$ element of a vector $x$ is denoted by $x_i$. 
Let $(\tilde{A}, \tilde{b})\in \mathbb{R}^{m\times n}\times \mathbb{R}^{m}$ denote random elements defined on a probability space $(\Xi,\mathcal{F},\mathbb{P})$ where $\Xi$ is the set of all possible realizations of the stochastic parameters, $\mathcal{F}$ is a $\sigma$-algebra of measurable events on $\Xi$, and $\mathbb{P}$ is a probability measure on $(\Xi,\mathcal{F})$.
Chance-constrained models provide a probabilistic framework for stochastic decision‐making by allowing controlled constraint violations. 
A joint chance-constrained program requires that all constraints hold simultaneously with probability at least~$1-\alpha$:
\begin{alignat*}{2}
    \min_{x \in \mathcal{X}} \quad & c^\top x \\
    \text{s.t.} \quad 
    & \mathbb{P}\big( \tilde{A}x \le \tilde{b} \big) \;\ge\; 1-\alpha,
\end{alignat*}
where $c\in \mathbb{R}^{n}$ is a cost vector, vector $x\in \mathcal{X}\subseteq \mathbb{Z}_{+}^k \times \mathbb{R}^{n-k}_{+}$ contains non-negative decision variables (with first $k$ variables being integer) and $\alpha \in [0,1]$ is the prescribed confidence level.  
Equivalently, the probability that any constraint is violated is limited by a risk budget:
\[
\mathbb{P}\big( \exists i:\, (\tilde{A}_{i\cdot} x > \tilde{b}_i) \big) \;\le\;  \alpha.
\]
 In this paper, we assume the probability law $\mathbb{P}$ is approximated by a finite empirical distribution supported on a finite set of sample scenarios. Thus, one can replace the chance constraints with finite-scenario approximations by sample-average approximation techniques, which can be formulated linearly using integer variables (e.g., see \cite{luedtke2008sample}, \cite[p. 351]{prekopa2013stochastic}).  

By contrast, an individual chance-constrained formulation enforces probabilistic feasibility separately for each constraint:
\begin{alignat*}{2}
    \min_{x \in \mathcal{X}} \quad & c^\top x \\
    \text{s.t.} \quad 
    & \mathbb{P}\big( \tilde{A}_{i\cdot} x \le \tilde{b}_i \big) \;\ge\; 1- \alpha_i, 
    \qquad i = 1,\dots,m,
\end{alignat*}
where the confidence level $\alpha_i$ may vary across constraints to represent heterogeneous reliability or safety requirements \cite{prekopa2013stochastic}.  
While the joint model enforces global reliability, the individual model allows constraint-wise flexibility at the expense of potentially weaker overall guarantees.  

The distinction between joint and individual chance-constrained formulations is central to understanding how language models interpret and encode probabilistic logic, specifically, whether they recognize collective feasibility events or treat constraints independently in mathematical syntax.

A two-stage deterministic mixed-integer linear programming (DMILP-2) formulation is a linear recourse model without uncertainty.  
First-stage variables $x$ are chosen initially; second-stage variables $y\in \mathcal{Y}\subseteq \mathbb{Z}_{+}^{p}\times \mathbb{R}^{r-p}_{+}$ adjust subsequently without uncertainty, and given $D \in \mathbb{R}^{z\times r}, B\in \mathbb{R}^{z\times n}, q\in \mathbb{R}^r$ and $d\in \mathbb{R}^{z}$, DMILP-2 is formulated as follows: 
\begin{alignat*}{2}
    \min_{x \in \mathcal{X}} \quad & c^\top x + h(x)\nonumber \\
    \text{s.t.} \quad & A x = b,
\end{alignat*}
where $h(x):=\min\limits_{y\in \mathcal{Y}}\{ q^\top y | D y = Bx + d\}$, represents the optimal cost of the second-stage response for a given first-stage decision $x$.  
This deterministic formulation serves as a baseline for comparing how LLMs can recover a hierarchical decision structure with random parameters when facing SMILP-2 problems.
The simplest SMILP-2 model generalizes the DMILP-2 formulation by allowing the second-stage data to be random. A scenario $\xi \in \Xi$ specifies a complete realization of the random second-stage problem data. We denote this data by the random vector $\Tilde{\xi} = \texttt{vec}(\Tilde{q}, \Tilde{D}, \Tilde{B}, \Tilde{d})$, where $\xi \in \Xi$ is a realization of $\Tilde{\xi}$,  $\Tilde{q}$ is a random cost vector for the recourse decision, $\Tilde{D}$ is a random constraint matrix for the recourse system, $\Tilde{B}$ is a random matrix that couples the first-stage decision to the second-stage system, and $\Tilde{d}$ is a random right-hand side vector with dimensions defined for their deterministic counterpart above. 

Given this setup, the SMILP-2 is
\begin{alignat}{2}
    \min_{x \in \mathcal{X}} \quad 
        & c^\top x + \mathbb{E}\!\left[h(x,\Tilde{\xi})\right] \label{eq:slp2} \\
    \text{s.t.} \quad 
        & A x = b, \nonumber 
\end{alignat}
where  $h(x,\Tilde{\xi}) := \min\limits_{y \in \mathcal{Y}}\{\Tilde{q}^\top y | \Tilde{D} y = \Tilde{B} x + \Tilde{d}\}.$ 
We assume relatively complete recourse and the $h(x,\tilde\xi)$ is integrable for all $x\in\mathcal{X}$.

The expectation $\mathbb{E}[h(x,\Tilde{\xi})]$ in~\eqref{eq:slp2} is taken with respect to the probability measure $\mathbb{P}$ on $(\Xi,\mathcal{F})$, i.e., $\mathbb{E}[h(x,\tilde\xi)] = \int_{\xi\in \Xi} h(x,\Tilde{\xi}))  \mathrm{d}\mathbb{P}(\xi),$ and represents the expected optimal recourse cost associated with $x$. 

In many practical settings, the uncertainty set $\Xi$ is modeled as a finite scenario set $\{\xi^1,\ldots,\xi^{s}\}$ with strictly positive scenario probabilities $p^1,\ldots,p^{s}$ satisfying $\sum_{i=1}^{s} p^i = 1$. In that case, each scenario $\xi^i$ induces deterministic data $\texttt{vec}(q^{i}, D^{i}, B^{i},d^{i})$, and the expectation operator in~\eqref{eq:slp2} reduces to a finite weighted sum. The equivalent extensive-form deterministic model is then given by
\begin{alignat*}{2}
    \min_{x\in \mathcal{X},\;y^i\in \mathcal{Y} (i=1,\dots, s)}& 
        && c^\top x + \sum_{i=1}^{s} p^{i} \,(q^{i})^\top y^{i} \\
    \text{s.t.}\ \qquad & 
        && A x = b, \\
      &  && D^{i} y^{i} = B^{i} x + d^{i}, \quad i = 1,\dots,s. 
\end{alignat*}
In this deterministic equivalent, the variable $y^{i}$ represents the recourse action that would be taken if scenario $\xi^{i}$ materializes. The linear system $D^{i} y^{i} = B^{i} x + d^{i}$ enforces scenario-wise feasibility, and the weighted sum $\sum_{i=1}^{s} p^{i} (q^{i})^\top y^{i}$ captures the expected recourse cost under the scenario probabilities. In the present context, it also provides a clean syntactic target for automatic model extraction and evaluation.

Next, we present prompts and the multi-agent framework.

\subsection{Design justification}
Our prompting methodology is deliberately hybrid because distinct prompt styles mitigate different, well-documented failure modes in NL-to-optimization translation and are supported by prior evidence, e.g., multi-agent prompting enhances reasoning \cite{hegazy2025diversitythoughtelicitsstronger, xiao2023chain} and mitigates extra-element generation \cite{huang2024mitigatingcatastrophicforgettinglarge}, and CoT improves LLMs' performance and reasoning capabilities \cite{sahoo2024systematic, wei2022chain, xiao2023chain}. We use the following methods: 

\emph{(i)}
CoT makes intermediate reasoning explicit, which is essential when expressing complex SO elements such as expectations, recourse mappings, and chance-constraints.
Empirically, CoT improves multi-step mathematical derivations \cite{sahoo2024systematic, wei2022chain, xiao2023chain}. Although we mentioned prompting methods in Section \ref{sec:lit} that slightly outperform CoT on deterministic optimization problems, CoT remains more practitioner-friendly. It is more interpretable and adaptable across problem settings without the need for external tools practiced by the recent works for SO problems (e.g., \cite{bertsimas2024robust}). 

\emph{(ii)}
Concise exemplars (sets $\rightarrow$ parameters $\rightarrow$ variables $\rightarrow$ objective/constraints) bias the model toward the correct algebraic skeleton, reducing notation drift and omitted elements.
Recent evaluations of LLMs report consistent gains from such structured prompting over pure zero-shot \cite{shojaee2025llm, tsoukalas2024putnambench, ying2024lean}.

\emph{(iii)}
We decompose the pipeline into extractor $\rightarrow$ formulator $\rightarrow$ reviewer $\rightarrow$ updater roles so that each stage can be checked locally (e.g., variables/sets before constraints; constraints).
Modular, agentic workflows have shown higher reliability than single-pass prompting in complex OR pipelines \cite{linzi2025systems, shojaee2025llm, tsoukalas2024putnambench}. In particular, agentic frameworks have been shown to exhibit greater robustness against forgetting and unsupported extra-element generation, as discussed in detail in Section~\ref{sec:agentic}.  

\emph{(iv)}
We request Python/Gurobi code (rather than textual math) to (a) demand explicit declarations of variables/constraints/objective, (b) avoid ambiguity in notation, (c) enable compile/runtime checks, and (d) gain better performance (see chain-of-code prompting in \cite{sahoo2024systematic}).
Code-based evaluation has proved to be easier than text-only outputs in optimization tasks \cite{Ramamonjison2022AugmentingOR, wang2022self}.

\emph{(v)}
Scenario expansion surfaces the block-angular structure and per-scenario feasibility that are hidden in compact SMILP-2 notation, making it possible to verify that (a) the expectation is correctly formed and (b) recourse constraints are indexed by scenario. Besides, this structure is easier to code as it is a linear model, which can help LLMs reduce compile errors.

\noindent
Collectively, these choices operationalize three principles: (a) make reasoning explicit (CoT/few-shot), (b) separate tasks and verify locally (multi-agent roles), and (c) evaluate in code and in algebra (Gurobi execution plus our soft-scoring metric).

\subsection{Multi-agent framework}\label{sec:agentic}

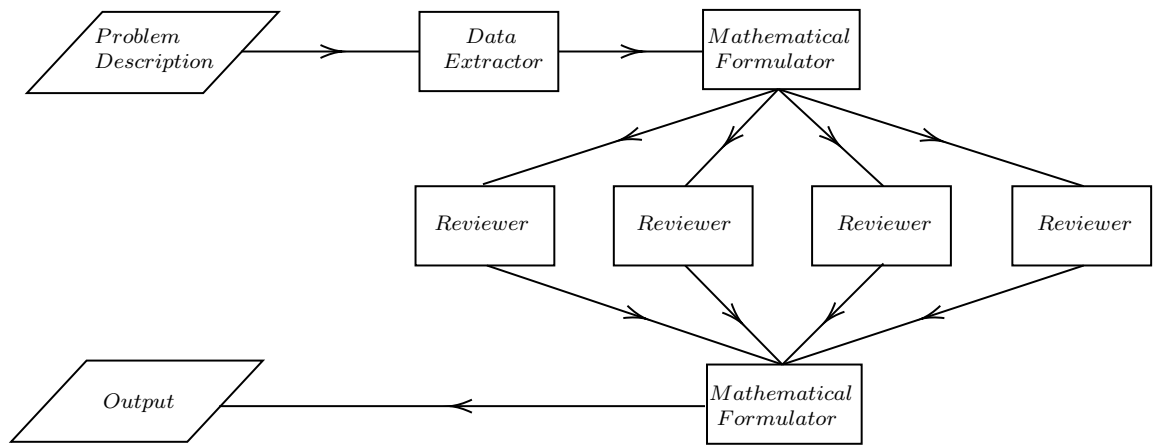
\begin{figure}[H]
\tikzset{every picture/.style={line width=0.75pt}} 
\begin{tikzpicture}[x=0.75pt,y=0.75pt,yscale=-1,xscale=1]

\draw   (102.1,12) -- (191,12) -- (152.9,53) -- (64,53) -- cycle ;

\draw   (262,12) -- (332,12) -- (332,52) -- (262,52) -- cycle ;

\draw   (405,11) -- (484,11) -- (484,51) -- (405,51) -- cycle ;

\draw   (260,100) -- (330,100) -- (330,140) -- (260,140) -- cycle ;

\draw   (360,100) -- (430,100) -- (430,140) -- (360,140) -- cycle ;

\draw   (460,100) -- (530,100) -- (530,140) -- (460,140) -- cycle ;

\draw   (561,100) -- (631,100) -- (631,140) -- (561,140) -- cycle ;

\draw   (407,190) -- (485,190) -- (485,230) -- (407,230) -- cycle ;

\draw    (443,51) -- (597,100) ;
\draw [shift={(525.72,77.32)}, rotate = 197.65] [color={rgb, 255:red, 0; green, 0; blue, 0 }  ][line width=0.75]    (10.93,-3.29) .. controls (6.95,-1.4) and (3.31,-0.3) .. (0,0) .. controls (3.31,0.3) and (6.95,1.4) .. (10.93,3.29)   ;

\draw    (443,51) -- (496,100) ;
\draw [shift={(473.91,79.57)}, rotate = 222.75] [color={rgb, 255:red, 0; green, 0; blue, 0 }  ][line width=0.75]    (10.93,-3.29) .. controls (6.95,-1.4) and (3.31,-0.3) .. (0,0) .. controls (3.31,0.3) and (6.95,1.4) .. (10.93,3.29)   ;
 
\draw    (443,51) -- (396,100) ;
\draw [shift={(415.35,79.83)}, rotate = 313.81] [color={rgb, 255:red, 0; green, 0; blue, 0 }  ][line width=0.75]    (10.93,-3.29) .. controls (6.95,-1.4) and (3.31,-0.3) .. (0,0) .. controls (3.31,0.3) and (6.95,1.4) .. (10.93,3.29)   ;

\draw    (443,51) -- (294,99) ;
\draw [shift={(362.79,76.84)}, rotate = 342.14] [color={rgb, 255:red, 0; green, 0; blue, 0 }  ][line width=0.75]    (10.93,-3.29) .. controls (6.95,-1.4) and (3.31,-0.3) .. (0,0) .. controls (3.31,0.3) and (6.95,1.4) .. (10.93,3.29)   ;

\draw    (597,140) -- (445,190) ;
\draw [shift={(515.3,166.87)}, rotate = 341.79] [color={rgb, 255:red, 0; green, 0; blue, 0 }  ][line width=0.75]    (10.93,-3.29) .. controls (6.95,-1.4) and (3.31,-0.3) .. (0,0) .. controls (3.31,0.3) and (6.95,1.4) .. (10.93,3.29)   ;

\draw    (496,139) -- (445,190) ;
\draw [shift={(466.26,168.74)}, rotate = 315] [color={rgb, 255:red, 0; green, 0; blue, 0 }  ][line width=0.75]    (10.93,-3.29) .. controls (6.95,-1.4) and (3.31,-0.3) .. (0,0) .. controls (3.31,0.3) and (6.95,1.4) .. (10.93,3.29)   ;

\draw    (396,140) -- (445,190) ;
\draw [shift={(424.7,169.29)}, rotate = 225.58] [color={rgb, 255:red, 0; green, 0; blue, 0 }  ][line width=0.75]    (10.93,-3.29) .. controls (6.95,-1.4) and (3.31,-0.3) .. (0,0) .. controls (3.31,0.3) and (6.95,1.4) .. (10.93,3.29)   ;

\draw    (296,140) -- (445,190) ;
\draw [shift={(376.19,166.91)}, rotate = 198.55] [color={rgb, 255:red, 0; green, 0; blue, 0 }  ][line width=0.75]    (10.93,-3.29) .. controls (6.95,-1.4) and (3.31,-0.3) .. (0,0) .. controls (3.31,0.3) and (6.95,1.4) .. (10.93,3.29)   ;
 
\draw   (94.1,188) -- (183,188) -- (144.9,229) -- (56,229) -- cycle ;

\draw    (406,211) -- (161,211) ;
\draw [shift={(277.5,211)}, rotate = 360] [color={rgb, 255:red, 0; green, 0; blue, 0 }  ][line width=0.75]    (10.93,-3.29) .. controls (6.95,-1.4) and (3.31,-0.3) .. (0,0) .. controls (3.31,0.3) and (6.95,1.4) .. (10.93,3.29)   ;

\draw    (172,32) -- (262,32) ;
\draw [shift={(223,32)}, rotate = 180] [color={rgb, 255:red, 0; green, 0; blue, 0 }  ][line width=0.75]    (10.93,-3.29) .. controls (6.95,-1.4) and (3.31,-0.3) .. (0,0) .. controls (3.31,0.3) and (6.95,1.4) .. (10.93,3.29)   ;

\draw    (332,32) -- (405,32) ;
\draw [shift={(374.5,32)}, rotate = 180] [color={rgb, 255:red, 0; green, 0; blue, 0 }  ][line width=0.75]    (10.93,-3.29) .. controls (6.95,-1.4) and (3.31,-0.3) .. (0,0) .. controls (3.31,0.3) and (6.95,1.4) .. (10.93,3.29)   ;

\draw (90,17.4) node [anchor=north west][inner sep=0.75pt]  [font=\scriptsize]  {$ \begin{array}{l}
Problem\ \\
Description
\end{array}$};

\draw (262,17.4) node [anchor=north west][inner sep=0.75pt]  [font=\scriptsize]  {$ \begin{array}{l}
\ \ \ \ Data\\
\ Extractor
\end{array}$};

\draw (399,17.4) node [anchor=north west][inner sep=0.75pt]  [font=\scriptsize]  {$ \begin{array}{l}
Mathematical\\
\ Formulator
\end{array}$};

\draw (268,113.4) node [anchor=north west][inner sep=0.75pt]  [font=\scriptsize]  {$Reviewer$};

\draw (370,113.4) node [anchor=north west][inner sep=0.75pt]  [font=\scriptsize]  {$Reviewer$};

\draw (471,113.4) node [anchor=north west][inner sep=0.75pt]  [font=\scriptsize]  {$Reviewer$};

\draw (572,113.4) node [anchor=north west][inner sep=0.75pt]  [font=\scriptsize]  {$Reviewer$};

\draw (400,197.4) node [anchor=north west][inner sep=0.75pt]  [font=\scriptsize]  {$ \begin{array}{l}
Mathematical\\
\ Formulator
\end{array}$};

\draw (101,203.4) node [anchor=north west][inner sep=0.75pt]  [font=\scriptsize]  {$Output$};

\end{tikzpicture}

\caption{Multi-agent prompting framework.}\label{fig: agents}
\end{figure}

To further utilize the reasoning abilities of LLMs, we designed a multi-agent framework. In this framework, distinct ChatGPT agents collaborate on solving parts of an SO problem. Each agent is prompted with tailored context and roles. Our multi-agent framework is depicted in Figure \ref{fig: agents}. Three important agents in this framework are the data extractor, the mathematical formulators, and the reviewers.
Data extractor elicits deterministic and stochastic variables and parameters together with sets and indices from the problem description and passes them to the next agent. The mathematical formulator receives the previous agent's outputs and problem description to code the model in Python (Gurobi execution). Reviewer agents evaluate the consistency and coherence of the output code from the mathematical formulator. They identify formulation issues or logical gaps. Even with identical prompts, four reviewer agents introduce stochastic diversity in reasoning paths and token sampling. The updating agent incorporates only those revisions justified by the reviewer feedback, closing a targeted feedback loop.

Agents' independent and collective analyses can reduce hallucinations, i.e., the generation of unsupported elements which are factually unsupported \cite{huang2024mitigatingcatastrophicforgettinglarge}  and improve reasoning accuracy \cite{hegazy2025diversitythoughtelicitsstronger}, and increase the likelihood that at least one agent identifies subtle algebraic or modeling errors overlooked by others. It also mitigates catastrophic forgetting, i.e., the loss of prior logical consistency during extended reasoning \cite{huang2024mitigatingcatastrophicforgettinglarge}, as agents can focus on specific tasks rather than the entire problem-solving process. Each agent has access to the problem description, which helps them be aware of the context and detect any potential errors caused by other agents, thereby mitigating the cascading errors phenomenon. \\ 

\subsection{Prompts}

In this section, we present the engineered prompts used in our experiments. We present three versions of CoT prompts, i.e., \texttt{cot\_s}, \texttt{cot\_s2}, and \texttt{cot\_instructions}. Our standard CoT prompt, \texttt{cot\_s}, is an adapted version of the CoT prompt taken from \cite{xiao2023chain}. We refer to it as standard since it is the simplest stochastic version of it in \cite{xiao2023chain}. We developed a sophisticated version of CoT, which includes additional hints, particularly for SMILP-2 problems. Finally, \texttt{cot\_instructions} includes instructions from Section \ref{sec:approach} for each of the stochastic problems. The simplest prompt is \texttt{standard\_s}, which is the adapted version of its deterministic counterpart in \cite{xiao2023chain}. It is called standard as it contains no hints or step-by-step reasoning procedure. Lastly, we present \texttt{agentic} prompt where each agent is specialized in a specific task. In what follows, we present the prompts with a brief discussion on them.

Prompts \ref{box:p32} and \ref{box:p33} refer to \texttt{standard\_s} and \texttt{cot\_s}. These are the simplest in this work, and we will compare the performance of the sophisticated prompts with the performance of these prompts.

\begin{promptboxenv}[box:p32]{Prompt 3.1: standard prompt for SO problems (\texttt{standard\_s}) adapted from \cite{xiao2023chain}}
You are a Python programmer in the field of Operations Research and Stochastic Optimization. Your proficiency in utilizing third-party libraries such as Gurobi is essential. In addition to your expertise in Gurobi, it would be great if you could also provide some background in related libraries or tools, like NumPy and SciPy.
You are given a specific problem. You aim to develop an efficient Python program that addresses the given problem.
Now the original problem is as follows:$\{\texttt{problem\_description}\}$. Here is a starter code:
$\{\texttt{\texttt{code\_example}}\}$. Give your Python code directly.

\end{promptboxenv}

\begin{promptboxenv}[box:p33]{Prompt 3.2: standard CoT prompt for SO problems adapted (\texttt{cot\_s}) from \cite{xiao2023chain}}
You are a Python programmer in the field of Stochastic Optimization. Your proficiency in utilizing third-party libraries such as Gurobi is essential. In addition to your expertise in Gurobi, it would be great if you could also provide some background in related libraries or tools, like NumPy and SciPy.
You are given a specific problem. You aim to develop an efficient Python program that addresses the given problem.
Now the original problem is as follows:
$\{\texttt{problem\_description}\}$.
Let's analyse the problem step by step, and then give your Python code directly.
Here is a starter code:
$\{\texttt{code\_example}\}$.

\end{promptboxenv}

The \texttt{\texttt{code\_example}} provided in the prompts refers to the code template. Providing a code template can give structure to the outputs generated by LLMs and pave the way for the evaluation of the GPT-generated codes.  
The advanced prompt \ref{box:p315} introduces more granular and structured steps with hints, making it worthwhile to investigate whether such precision leads to improved performance. For each problem category, it directs the model to decompose the description of the problem and extract sets and indices, parameters, stochastic and deterministic variables. In particular, for the SMILP-2 problems, it requires identifying first- and second-stage objectives and constraints and an extensive form of the problem.  
Prompt \ref{box:p3111} includes steps and hints of prompt \ref{box:p315} except that we have added instructions on the general structure of each type of problem category provided in Section \ref{sec:approach}. We hypothesize that these instructions may further help LLMs to focus on essential parts of the extraction of the elements and formulation.  
The multi-agent prompt is presented at \ref{box:p415}.  We hypothesize that this framework, beyond the known benefits of CoT, further reduces hallucinations by assigning each agent a narrowly tailored task and introducing a control loop that prevents cascading errors. 
Each prompt corresponds to a specific agent and is designed to operate independently; agents do not have access to one another's prompts.

A summary of the prompting methods and specifications is provided in Table \ref{tab:prompting_methods}. Note that prompting methods have been introduced in prior works. Following the structure of these methods, we generated SO-specific prompts. 

\begin{table}[H]
\caption{Prompting methods used in this paper with acronyms.}
\begin{tabular}{|l|c|c|c|c|c|}
\hline
\textbf{Prompt} & \textbf{Acronym} &\textbf{Instructions}&\textbf{Notes}\\
\hline
Zero-shot & \texttt{standard\_s} & \ding{55}& Stochasticity-aware version of \cite{xiao2023chain}\\
Multi-agent & \texttt{agentic} & \ding{55}& Explained in detail at \ref{sec:agentic}\\
CoT & \texttt{cot\_s} & \ding{55}&Stochasticity-aware version of \cite{xiao2023chain}\\
CoT& \texttt{cot\_s2}&\ding{55}& More precise guidance and steps than \texttt{cot\_s}\\
Few-shot  & \texttt{cot\_s\_instruction} & \ding{51}& Includes instructions\\
\hline
\end{tabular}
\label{tab:prompting_methods}
\end{table}

\begin{promptboxenv}[box:p315]{Prompt 3.3: Advanced CoT prompt for SO (\texttt{cot\_s2})}
$extract\_elements:$ You are an expert in Gurobi and Stochastic Optimization (SO). You will be given a verbal description of an SO problem. 
     It is either a two-stage stochastic mixed-integer linear programming (SMILP-2), its two-stage deterministic mixed-integer linear programming (DMILP-2) counterpart, or an individual or joint chance-constrained optimization problem. Your task is to identify and extract the 
     following:\\
      - Sets and indices (e.g., time periods, locations, resources, scenarios, etc.)\\
    - Deterministic and stochastic parameters (e.g., costs, capacities, probabilities, uncertain demand values, etc.)\\
    - Variables (e.g., first- and second- stage decisions in an SMILP-2 problem, integer variables, etc.)\\
     Present your Python code for the elements above. Here is a code template: $\{\texttt{code\_example}\}$. Your extraction will serve as the foundation for subsequent code implementation.\\
     $formulate\_model:$ Now, you are provided with the extracted components, including sets and indices, deterministic or stochastic parameters, and variables.
     Your task is to formulate the complete model in Python Gurobi. Specifically:\\
     Clearly define the objective function, including the first-stage costs and the expected recourse (second-stage) costs for SMILP-2 and DMILP-2 problems.\\
     Write all relevant constraints of the model. Use scenario-based indicator reformulations by integer variables and big-M to linearize chance constraints.\\
     $extensive\_form:$ Now, you are given the Python Gurobi formulation of the problem (including sets, indices, parameters, decision variables, objective function, and constraints).
     If it is an SMILP-2 or DMILP-2 problem, your task is to construct the extensive form of this problem in Python using Gurobi by modifying the previous code. Specifically:\\
     Enumerate all possible scenarios, associating each with its corresponding probability.\\
     Replace the expected value term in the objective function with scenario-specific expressions to form a deterministic equivalent if it is an SMILP-2 problem by replicating and customizing the second-stage constraints and variables for each scenario.\\
     Present the full model in a single-stage linear programming format suitable for direct input into a Gurobi solver in Python. Note the code example. \\
     Now the original problem is as follows:\\
     $\{\texttt{problem\_description}\}$.\\
     Let's analyse the problem step by step, then give your final Python code.
     Here is a starter code:
     $\{\texttt{code\_example}\}$.
\end{promptboxenv}

\begin{promptboxenv}[box:p3111]{Prompt 3.4: CoT with instructions (\texttt{cot\_instructions})}
\textit{extract\_elements}: You will be given a verbal description of a two-stage stochastic mixed-integer linear programming (SMILP-2) problem, or a two-stage deterministic mixed-integer linear programming (DMILP-2) problem, or an individual or joint chance-constrained problem. Your task is to identify and extract the following components from the text:\\
 - Sets and indices (e.g., time periods, locations, resources, scenarios, etc.)\\
- Deterministic and stochastic parameters (e.g., costs, capacities, probabilities, uncertain demand values, etc.)\\
- Variables (e.g., first- and second- stage decisions in an SMILP-2 problem, integer variables, etc.)\\
Present your answer in Python using Gurobi. 

The original problem is as follows:
\{\texttt{problem\_description}\}.

Let's analyse the problem step by step.
Here is a starter code:
\{\texttt{code\_example}\}.

Learn the following instructions to guide you further if it is an SMILP-2:\\
\{Instructions for SMILP-2\}.

If it is a chance-constraint model, use the following instructions: \\
\{Instructions for chance-constrained models\}.

If it is a DMILP-2 model, use the following instructions: \\
\{Instructions for DMILP-2 models\}.

\textit{formulate\_model}: Now, given the sets, indices, variables, and stochastic and deterministic parameters from the previous step,
Your task is to formulate the complete model. Specifically:\\
- Clearly define the objective function, including the first-stage objective and the expected recourse (second-stage) objective if it is an SMILP-2 problem.\\
- Code all constraints. Use scenario-based indicator reformulations by integer variables and big-M to linearize chance constraints.\\
\textit{extensive\_form} (only if it is an SMILP-2 problem): Now,
Your task is to construct the extensive form of this SMILP-2 model. Specifically:\\
- Enumerate all possible scenarios, associating each with its corresponding probability.\\
- Replace the expected value term in the objective function with scenario-specific expressions to form a deterministic equivalent by replicating and customizing the second-stage constraints and variables for each scenario.\\
- Present the full model in a single-stage linear programming format suitable for direct input into an LP solver.
Clearly label all variables and constraints by scenario.

Read the instructions below for further guidance on the extensive counterpart of an SMILP-2: \{Extensive form generation\}.

\textit{python\_code}: Now, you are provided with the complete mathematical formulation of the SO problem or DMILP-2 problem.

Your task is to generate the complete Python code using Gurobi solver. 
\end{promptboxenv}

\begin{promptboxenv}[box:p415]{Prompt 3.5: multi-agent prompt for SO (\texttt{agentic})}
$Data\ extractor\ agent:$
You will be given a verbal description of a Stochastic Optimization problem, such as a two-stage stochastic mixed-integer linear programming (SMILP-2) problem (or its two-stage deterministic mixed-integer linear programming counterpart)  or individual and joint chance-constrained problems. Your task is to identify and extract the 
             following components from the text:\\
              - Sets and indices (e.g., time periods, locations, resources, scenarios, etc.)\\
             - Deterministic and stochastic parameters (e.g., costs, capacities, probabilities, uncertain demand values, etc.)\\
             - Variables (e.g., first- and second- stage decisions in an SMILP-2 problem, integer variables, etc.)\\
             Present your answer in Python code using Gurobi. Below is also a code template, and 
              problem description:\\
            $\{\texttt{problem\_description}\}$ + $\{\texttt{code\_example}\}$.\\
$Formulation\ agent:$ You will be given a verbal description of a Stochastic Optimization problem, such as a two-stage stochastic mixed-integer linear programming (SMILP-2) problem (or its two-stage deterministic mixed-integer programming counterpart), individual or joint chance-constrained problems, along with the extracted components 
                        including sets, indices, stochastic and deterministic parameters, 
                        and variables.
                        Your task is to code the complete model in Python using Gurobi. Specifically:\\
                        Clearly define the objective function.\\
                        Code all relevant constraints. Specifically, if it is an SMILP-2 problem, you need to code the extensive form of the problem. Use scenario-based indicator reformulations by integer variables and big-M to linearize chance constraints.\\
                        New problem description:\\
                        $\{\texttt{problem\_description}\}$.\\
                        Below is the extraction output:\\
                       $ \{extraction\_output\}$.\\
                        Below is the code template you should follow:\\
                        $\{\texttt{code\_example}\}$.\\
                $Reviewer\ agent:$
                You are a reviewer agent specialized in Stochastic Optimization problems. 
                        You are provided with a problem description and a final Python code using Gurobi solver. 
                        Review them carefully for potential mistakes (such as variables, objective function, constraints, parameters, and syntax errors) and any additional elements that might be required. 
                        Provide concise and precise feedback.\\
                        Problem description:\\
                        $\{\texttt{problem\_description}\}$.\\
                        Mathematical formulation:\\
                        $\{code\_formulation\_output\_1\}$.\\
                        $Formulation\ agent\ 2:$\\
                    You are a specialized updating agent in Stochastic Optimization.
                    You are provided with the following information:\\
                     Problem description:\\
                  $  \{\texttt{problem\_description}\}$.\\
                     Current code:\\
                    $\{current\_code\}$.\\
                    Feedback from reviewer agents:\\
                    $\{reviewers\_feedback\}$.\\
                    Review the feedback carefully. If the feedback indicates valid improvements, update the code accordingly.
                    Return the updated final code. 
                    Do not include any additional text.
\end{promptboxenv}

\subsection{Motivating examples}

We provide several examples where text output is desired. Below is an example of a joint chance-constrained problem.

\begin{tcolorbox}[promptbox, title={Prompt: joint chance-constrained, adapted from \cite{asadpour2020online, chen2023onlinealgorithmchanceconstrained}}]
You are an expert in analyzing joint chance-constrained models. Here is our problem. 
You are managing a resource allocation problem. You have to allocate amounts to two projects. The actual resource needs are uncertain, and you want to ensure both project demands are met simultaneously with 95\% probability.  

Read the following instructions to guide you and give the mathematical formulation.
\{Prompt: Instructions for joint chance-constrained models\}
\end{tcolorbox}

\begin{tcolorbox}[promptbox, title={ChatGPT-3.5-Turbo's response for the joint chance-constrained modeling:}]
Joint Chance-Constrained Resource Allocation
Decision Variables:
\begin{itemize}
  \item $x_1, x_2 \in \mathbb{R}_+$: amount of resource allocated to Project 1 and Project 2
\end{itemize}
Random Variables:
\begin{itemize}
  \item $\tilde{d}_1, \tilde{d}_2$: random demands for Project 1 and Project 2
\end{itemize}
Model:
\begin{alignat*}{2}
    &\min_{x_1, x_2} \quad && x_1 + x_2 \\
    &\text{s.t.} \quad && \mathbb{P}\left( 
      \begin{bmatrix}
      x_1 \geq \tilde{d}_1 \\
      x_2 \geq \tilde{d}_2
      \end{bmatrix}
    \right) \geq 0.95 \\
    &&& x_1 \geq 0,\quad x_2 \geq 0
\end{alignat*}
\end{tcolorbox}

The strengths of the solution are that the model correctly interprets the requirement as a joint chance-constrained, and the response provides a separation between decision variables and random parameters.

The following example is for an individual chance-constrained model.

\begin{tcolorbox}[promptbox, title={Prompt: individual chance-constrained, adapted from \cite{asadpour2020online}}]
You are an expert in analyzing individual chance-constrained models. Here is our problem: 
You are responsible for allocating delivery trucks from two warehouses to fulfill uncertain demands at two retail stores, Store A and Store B. The number of trucks allocated to each store is denoted by \( x_1 \) and \( x_2 \), respectively. However, the exact demand at each store is not known in advance and is modeled as a normal random variable:
\begin{itemize}
    \item \( \tilde{d}_1 \sim \mathcal{N}(100, 10^2) \) for Store A,
    \item \( \tilde{d}_2 \sim \mathcal{N}(150, 15^2) \) for Store B.
\end{itemize}
Each truck delivers exactly one unit of goods. The goal is to minimize the total number of trucks dispatched, while ensuring that the supply meets the demand at each store with high probability:
\begin{itemize}
    \item Store A should receive enough goods with at least 95\% probability.
    \item Store B should receive enough goods with at least 90\% probability.
\end{itemize} 
Read the following instructions to guide you and give the mathematical formulation.
\{Prompt: Instructions for individual chance-constrained models\}
\end{tcolorbox}

\begin{tcolorbox}[promptbox, title={ChatGPT-4's response for the individual chance-constrained modeling:}]
Mathematical Formulation: Individual Chance-Constrained Model
\begin{alignat*}{2}
    &\min_{x_1, x_2} \quad && x_1 + x_2 \\
    &\text{s.t.} \quad 
        &&\mathbb{P}\left( x_1 \geq \tilde{d}_1 \right) \geq 0.95 \quad \text{(Store A)} \\
    &&&\mathbb{P}\left( x_2 \geq \tilde{d}_2 \right) \geq 0.90 \quad \text{(Store B)} \\
    &&&x_1 \geq 0,\quad x_2 \geq 0
\end{alignat*}
\end{tcolorbox}

The solution's strengths are that the model correctly interprets the requirements as individual chance-constrained, the objective is consistent with the problem statement, and the formulation is compact and clearly structured. The weakness is that trucks are indivisible units; $x_1$ and $x_2$ should be modeled as integers.

The following example is for an SMILP-2 model. 

\begin{tcolorbox}[promptbox, title={Prompt: SMILP-2, adapted from \cite{maggioni2017scenario}}]
You are an expert in analyzing two-stage stochastic mixed-integer linear programming models (SMILP-2). Here is our problem: 
We consider a supply-planning problem in which a producer books transportation capacity in advance to satisfy uncertain factory demand. After demand is realized, unused vehicle bookings may be cancelled at a fee, and any remaining shortage can be covered by purchasing from an external supplier at a higher, uncertain cost. The goal is to determine how many vehicles to book from each plant to minimize the expected total cost. 

Read the following instructions to guide you and give the mathematical formulation.
\{Prompt: Instructions for SMILP-2 models\}
\end{tcolorbox}

\begin{tcolorbox}[promptbox, title={ChatGPT-4's response for the SMILP-2 modeling:}] 
\begin{alignat*}{2}
\text{Sets:} \quad
& I  \text{: set of origins (plants)} \\
& \Omega  \text{: set of scenarios, with probability } p_\omega \text{ for } \omega \in \Omega \\[0.5em]
\text{Parameters:} \quad
& d^\omega  \text{: demand at factory under scenario } \omega \\
& c_i  \text{: cost to book one vehicle from plant } i \\
& \bar{c}_i^\omega  \text{: cost of actually using a booked vehicle from } i \text{ in } \omega \\
& \gamma_i  \text{: cancellation fee per unused vehicle at plant } i \\
& q^\omega  \text{: unit cost to purchase shortage from external supplier in } \omega \\
& Q  \text{: vehicle capacity (fixed, full-load)} \\[0.5em]
\text{First-stage decision:} \quad
& x_i  \text{: number of vehicles to book from plant } i \\[0.5em]
\text{Second-stage decisions:} \quad
& y_i^\omega  \text{: number of booked vehicles used from } i \text{ under } \omega \\
& z^\omega  \text{: quantity purchased from external supplier under } \omega \\[0.5em]
\min_{x} \quad  \sum_{i \in I} c_i x_i + \mathbb{E}_\omega\left[\phi(x,\omega)\right] \\
\text{s.t.} \quad  x_i \in \mathbb{Z}_+ \quad \forall i \in I \\[0.5em]
\text{where} \quad 
\phi(x,\omega) := \min_{y^\omega, z^\omega} \ 
& \sum_{i \in I} \left( \bar{c}_i^\omega y_i^\omega + \gamma_i (x_i - y_i^\omega)_+ \right) + q^\omega z^\omega \\
\text{s.t.} \quad 
& \sum_{i \in I} Q y_i^\omega + z^\omega \ge d^\omega \\
& 0 \le y_i^\omega \le x_i, \forall i \in I, \quad z^\omega \ge 0
\end{alignat*}

\end{tcolorbox}

The strengths of the solution are that the formulation correctly captures the two-stage structure, the model explicitly links first- and second-stage decisions through the constraints $0 \le y_i^\omega \le x_i$, and the recourse function is well-structured. 
The weaknesses are that the recourse cost function is not linearized, and the demand-satisfaction constraint assumes continuous or fractional vehicle usage.

The prompt below is for the DMILP-2 problem.

\begin{tcolorbox}[promptbox, title={Prompt: DMILP-2 problem, adapted from \cite{asadpour2020online}}]
You are an expert in analyzing two-stage deterministic mixed-integer linear programming (DMILP-2) models. Here is our problem: 
You're managing inventory for two warehouses. In the first stage, you decide how much product to order for each warehouse. In the second stage, once customer demands from two regions are revealed (and they’re deterministic), you decide how to ship products from the warehouses to meet those demands. The objective is to minimize total ordering and shipping costs while satisfying all customer demand and not exceeding warehouse inventory levels.
Read the following instructions to guide you and give the mathematical formulation. 
\{Prompt: Instructions for DMILP-2 models\}
\end{tcolorbox}

\begin{tcolorbox}[promptbox, title={ChatGPT-4's response for the DMILP-2 modeling:}]
 The goal is to minimize ordering and shipping costs.
\begin{alignat}{2}
    \min_{x}\quad & c^\top x + h(x) \\ 
    \text{s.t.} \nonumber\quad
    & x \geq 0 \\
    \text{where}\nonumber\\
    &h(x) = 
    \begin{aligned}
        \min_{y}\quad & \sum_{i=1}^2 \sum_{j=1}^2 s_{ij} y_{ij} \\
        \text{s.t.} \quad 
        & \sum_{j=1}^2 y_{ij} \leq x_i &&\quad \text{(do not exceed warehouse stock)} \\
        & \sum_{i=1}^2 y_{ij} = d_j &&\quad \text{(meet regional demand)} \\
        & y_{ij} \geq 0 &&\quad \forall i,j
    \end{aligned}
\end{alignat}
where  \( x_i \): quantity ordered for warehouse \( i \) (first-stage decision), \( y_{ij} \): quantity shipped from warehouse \( i \) to region \( j \) (second-stage decision), \( c_i \): cost per unit ordered at warehouse \( i \), \( s_{ij} \): shipping cost per unit from warehouse \( i \) to region \( j \) and \( d_j \): known demand in region \( j \)

\end{tcolorbox}

The strength of the solution is that it correctly identifies and applies a DMILP-2 structure. The weakness is in the declaration of index sets for constraints. 

These observations motivate us to develop a scoring system to automate the evaluation and quantification of solutions at scale.

\subsection{Scoring}

The scoring system evaluates ChatGPT’s generated model's Python code based on output code correctness and structural matching to a ground-truth code. Exact matches receive full credit, while variables, constraints, and objectives are compared using a matching strategy that assigns partial credit and reports any runtime or compile-time errors.
The scoring system’s operations are illustrated by Algorithm \ref{alg:11}. The indicator function used in this paper is defined as 
$
\mathbf{1}\{P\}= 
\begin{cases}
1, & \text{if } P \text{ is true},\\
0, & \text{otherwise}.
\end{cases}
$, and $|.|$ indicates the cardinality of a set.

\begin{algorithm}[H] 
\small 
\caption{Scoring system}\label{alg:11}

\begin{algorithmic}[1]
\Require \texttt{true\_code}, \texttt{generated\_code}
\Ensure $(P_{\mathrm{total}},P_{\mathrm{extra}},P_{\mathrm{var}},P_{\mathrm{cons}},P_{\mathrm{obj}})\in[0,100]^5$

\State \textbf{If} running generated code on a sample equals true optimal objective value \textbf{then return} accuracy $=100$.

\State Parse variables $V^{\mathrm{T}},V^{\mathrm{G}}$ via \texttt{.addVars}, with arg strings $\mathcal{A}^{\mathrm{T}},\mathcal{A}^{\mathrm{G}}$.
\State Parse constraints $C^{\mathrm{T}},C^{\mathrm{G}}$ via first arg of \texttt{.addConstr(s)(·)} (drop outer \texttt{quicksum}).
\State Parse objective $O^{\mathrm{T}},O^{\mathrm{G}}$ via \texttt{.setObjective(·,GRB.(MAX)MINIMIZE)}; inline if a bare name; drop outer sums.

\State Define normalizers:
\[
\mathcal{N}_{\mathrm{idx}}:\ [\cdot]\!\mapsto\![\mathrm{idx}],\quad
\mathcal{N}_{\mathrm{var}}(s;\upsilon): \upsilon\!\mapsto\!\texttt{\_X\_},\ \text{others}\mapsto\texttt{\_Yk\_}, K=0,\cdots, k,
\]
\[
\mathcal{R}(r)=
\begin{cases}
\texttt{lhs}-\texttt{rhs},& r:\ \texttt{lhs}\le\texttt{rhs},\\
\texttt{rhs}-\texttt{lhs},& r:\ \texttt{lhs}\ge\texttt{rhs},\\
\texttt{lhs}-\texttt{rhs},& r:\ \texttt{lhs}=\texttt{rhs},\\
r,& \text{otherwise}.
\end{cases}
\]

\State Define $\mathrm{occ}$$(s,\upsilon)$ as the count of occurrences of $\upsilon$ plus occurrences of $\upsilon[\cdot]$.

\State Define SymPy equivalence \textsc{Eq}$(a,b)$ as:
replace each \texttt{sum/quicksum} sub-expression by a canonical symbol determined by its normalized body and sorted generator clauses, then equivalence is true if \textsc{Eq}$(a,b) =0$.

\For{$g\in V^{\mathrm{G}}$}\For{$t\in V^{\mathrm{T}}$} \label{line:pair}
  \State $S(g,t)\gets 0$
  \State $S(g,t)\ \!+\!=\!|\{k\in\{\texttt{lb},\texttt{ub},\texttt{vtype}\}:\ k=\cdot \text{ appears identically in }\mathcal{A}^{\mathrm{G}}(g),\mathcal{A}^{\mathrm{T}}(t)\}|$ \hfill (args)
   \State $S(g,t)\ \!+\!=\min\ \{\sum_{s\in C^{\mathrm{T}}\cup\{O^{\mathrm{T}}\}}\!\!\mathrm{occ}(s,t),\ \sum_{s\in C^{\mathrm{G}}\cup\{O^{\mathrm{G}}\}}\!\!\mathrm{occ}(s,g)\}$ \hfill (frequency)
   \For{$c^{\mathrm{T}}\in C^{\mathrm{T}}$ \textbf{such that} $t$ occurs in $c^{\mathrm{T}}$} \hfill (constraints)
    \For{$c^{\mathrm{G}}\in C^{\mathrm{G}}$ \textbf{such that} $g$ occurs in $c^{\mathrm{G}}$}
    \State $\tilde c^{\mathrm{T}}\!\gets\!\mathcal{N}_{\mathrm{var}}(\mathcal{N}_{\mathrm{idx}}(c^{\mathrm{T}});t)$,\ \ $\tilde c^{\mathrm{G}}\!\gets\!\mathcal{N}_{\mathrm{var}}(\mathcal{N}_{\mathrm{idx}}(c^{\mathrm{G}});g)$
    \State $S(g,t)\ \!+\!=2\cdot \mathbf{1}\{\tilde c^{\mathrm{T}}\!=\!\tilde c^{\mathrm{G}}\}+2\cdot\mathbf{1}\{\textsc{Eq}(\mathcal{R}(\tilde c^{\mathrm{T}}),\mathcal{R}(\tilde c^{\mathrm{G}}))\}+2\cdot \mathbf{1}\{\rm rmsp(\tilde c^{\mathrm{T}})\!=\!\rm rmsp(\tilde c^{\mathrm{G}})\}$
    \State (\texttt{rmsp} is a normalization function deleting whitespace and punctuation)
  \EndFor\EndFor
  \If{$t\in O^{\mathrm{T}}$ and $g\in O^{\mathrm{G}}$}  \hfill (objective)
    \State $\tilde O^{\mathrm{T}}\!\gets\!\mathcal{N}_{\mathrm{var}}(\mathcal{N}_{\mathrm{idx}}(O^{\mathrm{T}});t)$,\ \ $\tilde O^{\mathrm{G}}\!\gets\!\mathcal{N}_{\mathrm{var}}(\mathcal{N}_{\mathrm{idx}}(O^{\mathrm{G}});g)$
    \State $S(g,t)\ \!+\!=2\cdot \mathbf{1}\{\tilde O^{\mathrm{T}}\!=\!\tilde O^{\mathrm{G}}\}+2\cdot \mathbf{1}\{\textsc{Eq}(\mathcal{R}(\tilde O^{\mathrm{T}}),\mathcal{R}(\tilde O^{\mathrm{G}}))\}+2\cdot \mathbf{1}\{\rm rmsp(\tilde O^{\mathrm{T}})\!=\!\rm rmsp(\tilde O^{\mathrm{G}})\}$
    \State $S(g,t)\ \!+\!=2\cdot|\,\{\text{coef}\cdot t[\cdot]\!\subset\!O^{\mathrm{T}}\}\cap\{\text{coef}\cdot g[\cdot]\!\subset\!O^{\mathrm{G}}\}\,|$ \hfill (term align)
  \EndIf
\EndFor\EndFor

\State Build a greedy partial one-to-one mapping $\mu:V^{\mathrm{G}}\rightharpoonup V^{\mathrm{T}}$ by selecting pairs $(g,t)$ in descending $S(g,t)$ without reuse; ignore $S(g,t)=0$.
\State Rename whole-word $g\mapsto\mu(g)$ in \texttt{generated\_code} (longest-first) to obtain \texttt{aligned\_generated\_code}.
\State Re-parse $\widehat V^{\mathrm{G}},\widehat C^{\mathrm{G}},\widehat O^{\mathrm{G}}$ from \texttt{aligned\_generated\_code}.

\State $M_{\mathrm{var}}\!\gets |V^{\mathrm{T}}\cap \widehat V^{\mathrm{G}}|$.\\
$M_{\mathrm{cons}}\gets \left|\left\{c^{\mathrm{T}}\in C^{\mathrm{T}}:\ \exists \hat c\in\widehat C^{\mathrm{G}}\ \text{s.t.}\ 
\mathcal{N}_{\mathrm{idx}}(c^{\mathrm{T}})=\mathcal{N}_{\mathrm{idx}}(\hat c)\ \text{or algebraic/structural equality holds after applying }\mu\right\}\right|$.
\State $M_{\mathrm{obj}}\gets \mathbf{1}\!\left[\exists \hat O\in\{\widehat O^{\mathrm{G}}\}\ \text{s.t. } \mathcal{N}_{\mathrm{idx}}(O^{\mathrm{T}})=\mathcal{N}_{\mathrm{idx}}(\hat O)\ \text{or algebraic/structural equality holds}\right]$.

\State $T_{\mathrm{var}}\!\gets|V^{\mathrm{T}}|$, $T_{\mathrm{cons}}\!\gets|C^{\mathrm{T}}|$, $T_{\mathrm{obj}}\!\gets |O^{\mathrm{T}}|$
\State $P_{\mathrm{var}}\!=\!100\cdot\frac{M_{\mathrm{var}}}{T_{\mathrm{var}}}$,\quad
$P_{\mathrm{cons}}\!=\!100 \cdot \frac{M_{\mathrm{cons}}}{T_{\mathrm{cons}}}$,\quad
$P_{\mathrm{obj}}\!=\!100\cdot\frac{M_{\mathrm{obj}}}{T_{\mathrm{obj}}}$
\State $P_{\mathrm{total}}\!=\!100\cdot\frac{M_{\mathrm{var}}+M_{\mathrm{cons}}+M_{\mathrm{obj}}}{T_{\mathrm{var}}+T_{\mathrm{cons}}+T_{\mathrm{obj}}}$
\State $P_{\mathrm{extra}}\!=\!100\cdot\frac{E}{G}$ where $G\!=\!|\widehat V^{\mathrm{G}}|+|\widehat C^{\mathrm{G}}|+|\widehat O^{\mathrm{G}}|$ and $E$ is number of unmatched generated components.
\State \Return $(P_{\mathrm{total}},P_{\mathrm{extra}},P_{\mathrm{var}},P_{\mathrm{cons}},P_{\mathrm{obj}})$
\end{algorithmic}
\end{algorithm}

\subsection{Formal specification of the scoring metric}
\label{sec:scoring-metric}

We evaluate a candidate (generated) model's code against a reference (true) model's code by parsing their Gurobi code strings and computing matching scores. While a matching optimal objective solution suggests a high-quality model, it may still lack essential variables or constraints. To prevent false positives, we can investigate matching scores for variables, constraints, and the objective function.

\textit{Inputs and outputs:}
Let \texttt{true\_code} and \texttt{generated\_code} be the two input strings. The procedure returns the following percentages: $P_{\mathrm{var}},\; P_{\mathrm{cons}},\; P_{\mathrm{obj}},\; P_{\mathrm{total}},\; P_{\mathrm{extra}} \in [0,100].$

If a preliminary execution of the generated code on a sample instance produces output exactly equal to the ground-truth output (optimal objective solution), the procedure returns \textit{Accuracy} $=100$. Yet, we check for structural matching to alarm false positives. 

\textit{Parsed components:}
From each code string, we extract:
\begin{itemize}
\item Variables: every assignment $v = \texttt{model.addVars}(\mathrm{args})$
yields a variable name $v$ and its raw argument string $\mathrm{args}$ (which may contain \texttt{lb=}\,$\cdot$, \texttt{ub=}\,$\cdot$, \texttt{vtype=}\,$\cdot$). Let $V^{\mathrm{T}}$ (respectively \ $V^{\mathrm{G}}$) be the set of true (respectively\ generated) variable names, with argument strings $\mathrm{args}^{\mathrm{T}}_t$ for $t\in V^{\mathrm{T}}$ and $\mathrm{args}^{\mathrm{G}}_g$ for $g\in V^{\mathrm{G}}$.

\item Constraints: from every call \texttt{model.addConstr(s)(}\,$\cdot$\,\texttt{)}, we take its first argument as a single constraint string (e.g., \texttt{a[i]*x[i] <= b[i]}). We denote these sets by $C^{\mathrm{T}}$ and $C^{\mathrm{G}}$.

\item Objective function: from \texttt{model.setObjective(}\,$\cdot$\,\texttt{, GRB.(MAX)MINIMIZE)} we extract the first argument as a single expression $O^{\mathrm{T}}$ or $O^{\mathrm{G}}$. If the argument is a bare identifier (e.g., \texttt{obj}), its right-hand side is inlined from its assignment. For example, if the output is \texttt{obj = gp.quicksum(cost[i] * x[i] for i in I)} and \texttt{model.setObjective(obj, GRB.MINIMIZE)}, the expression for objective function is \texttt{cost[i] * x[i]}. 
Outermost wrappers of the form \texttt{quicksum(\,.\,)} or \texttt{sum(\,.\,)} are dropped prior to comparison.
\end{itemize}

 The implementation considers \texttt{addVars}, \texttt{addConstr}, and \texttt{setObjective(\,$\cdot$, GRB.MINIMIZE)} only.

\textit{Normalizations and equivalence:}
We use three normalizations:
\begin{enumerate}
\item Index normalization, $\mathcal{N}_{\mathrm{idx}}$: we replace any bracketed index by a placeholder,
\(
[\cdot] \mapsto [\texttt{idx}],
\)
and normalize simple generator phrases (e.g., \texttt{for k in range(…)}).
\item Variable-centric structural normalization, $\mathcal{N}_{\mathrm{var}}(\,\cdot\,;\upsilon)$: given a target symbol $\upsilon$, we map $\upsilon\!\mapsto\!\texttt{\_X\_}$ and every other symbol to placeholders \texttt{\_Y0\_}, \texttt{\_Y1\_}, \dots\ deterministically by first occurrence.
\item Relational residualization, $\mathcal{R}(\cdot)$: for a string $r$ of the form \texttt{lhs <= rhs}, \texttt{lhs >= rhs}, or \texttt{lhs == rhs}, we define
\[
\mathcal{R}(r)=
\begin{cases}
\texttt{lhs}-\texttt{rhs},& r:\ \texttt{lhs}\le\texttt{rhs},\\
\texttt{rhs}-\texttt{lhs},& r:\ \texttt{lhs}\ge\texttt{rhs},\\
\texttt{lhs}-\texttt{rhs},& r:\ \texttt{lhs}=\texttt{rhs},\\
r,& \text{otherwise}.
\end{cases}
\]

\end{enumerate}
Using SymPy’s \texttt{sympify} and \texttt{simplify}, we declare:
\begin{align*}
\text{(structural equivalence)}\quad
&\mathcal{N}_{\mathrm{var}}\!\bigl(\mathcal{N}_{\mathrm{idx}}(s_1);\upsilon_1\bigr)
=
\mathcal{N}_{\mathrm{var}}\!\bigl(\mathcal{N}_{\mathrm{idx}}(s_2);\upsilon_2\bigr),\\
\text{(algebraic equivalence)}\quad
&\text{$\mathcal{R}\!\bigl(\mathcal{N}_{\mathrm{var}}(\mathcal{N}_{\mathrm{idx}}(s_1);\upsilon_1)\bigr)$ and
$\mathcal{R}\!\bigl(\mathcal{N}_{\mathrm{var}}(\mathcal{N}_{\mathrm{idx}}(s_2);\upsilon_2)\bigr)$ simplify to equal forms.}
\end{align*}
We also use whitespace-insensitive equality as a tie-breaker on normalized strings.

\textit{Per-variable pair score:}
For each $(g,t)\!\in\! V^{\mathrm{G}}\!\times\! V^{\mathrm{T}}$ we compute following score:
$$
S(g,t) \;=\; S_{\mathrm{var}}(g,t)+S_{\mathrm{cons}}(g,t)+S_{\mathrm{obj}}(g,t)+S_{\mathrm{freq}}(g,t)+S_{\mathrm{term}}(g,t),
$$
With the following components:
\begin{enumerate}
\item Argument match, $S_{\mathrm{var}}\!\in\!\{0,1,2,3\}$: for each $k\in\{\texttt{lb},\texttt{ub},\texttt{vtype}\}$, we add $1$ if the sub-string \texttt{k} appears in both $\mathrm{args}^{\mathrm{G}}_g$ and $\mathrm{args}^{\mathrm{T}}_t$ and these sub-strings are identical.

\item Constraint structure, $S_{\mathrm{cons}}\!\in\!\mathbb{Z}_{\ge 0}$: for every $c^{\mathrm{T}}\!\in\!C^{\mathrm{T}}$ containing $t$ and every $c^{\mathrm{G}}\!\in\!C^{\mathrm{G}}$ containing $g$, we set
\(
\tilde c^{\mathrm{T}}=\mathcal{N}_{\mathrm{var}}(\mathcal{N}_{\mathrm{idx}}(c^{\mathrm{T}});t),\;
\tilde c^{\mathrm{G}}=\mathcal{N}_{\mathrm{var}}(\mathcal{N}_{\mathrm{idx}}(c^{\mathrm{G}});g).
\)
Then, we add
$$
\begin{cases}
+2 & \text{if } \tilde c^{\mathrm{T}}=\tilde c^{\mathrm{G}},\\
+2 & \text{if } \mathcal{R}(\tilde c^{\mathrm{T}}) \text{ and } \mathcal{R}(\tilde c^{\mathrm{G}}) \text{ simplify to equal forms},\\
+2 & \text{if } \text{rmsp}(\tilde c^{\mathrm{T}})=\text{rmsp}(\tilde c^{\mathrm{G}}).
\end{cases}
$$
(All three checks are performed and accumulated independently. The operator rmsp(.) eliminates white-spaces).

\item Objective structure, $S_{\mathrm{obj}}\!\in\!\{0,2,4,6\}$: if $t$ occurs in $O^{\mathrm{T}}$ and $g$ occurs in $O^{\mathrm{G}}$, we define
$$
\tilde O^{\mathrm{T}}=\mathcal{N}_{\mathrm{var}}(\mathcal{N}_{\mathrm{idx}}(O^{\mathrm{T}});t),\;
\tilde O^{\mathrm{G}}=\mathcal{N}_{\mathrm{var}}(\mathcal{N}_{\mathrm{idx}}(O^{\mathrm{G}});g).
$$
Then, we add
$$
\begin{cases}
+2 & \text{if } \tilde O^{\mathrm{T}}=\tilde O^{\mathrm{G}},\\
+2 & \text{if } \mathcal{R}(\tilde O^{\mathrm{T}})\text{ and } \mathcal{R}(\tilde O^{\mathrm{G}}) \text{ simplify to equal forms},\\
+2 & \text{if } \text{remove-spaces}(\tilde O^{\mathrm{T}})=\text{remove-spaces}(\tilde O^{\mathrm{G}}).
\end{cases}
$$

\item Usage frequency, $S_{\mathrm{freq}}\!\in\!\mathbb{Z}_{\ge 0}$: let $\mathrm{occ}(s,\upsilon)$ be the exact token count of $\upsilon$ (bare or indexed) in a string $s$. Then
$$
S_{\mathrm{freq}}(g,t)=
\min \ \!\Bigl\{\sum_{s\in C^{\mathrm{T}}\cup O^{\mathrm{T}}}\!\!\mathrm{occ}(s,t),\;
\sum_{s\in C^{\mathrm{G}}\cup O^{\mathrm{G}}}\!\!\mathrm{occ}(s,g)\Bigr\}.
$$

\item Term matching, $S_{\mathrm{term}}\!\in\!\mathbb{Z}_{\ge 0}$: we define
$$
T^{\mathrm{T}}(t)=\{\text{“coef}*\;t[\cdot]\text{” subterms in }O^{\mathrm{T}}\}
$$
and
$$
T^{\mathrm{G}}(g)=\{\text{“coef}*\;g[\cdot]\text{” subterms in }O^{\mathrm{G}}\}.
$$
Then
$$
S_{\mathrm{term}}(g,t)=2\,|\,T^{\mathrm{T}}(t)\cap T^{\mathrm{G}}(g)\,|.
$$

\end{enumerate}

\textit{Variable matching and renaming:}
we form all pairs $(g,t)$ with scores $S(g,t)$ and sort them in descending order. Then, we build a partial one-to-one mapping $\mu:V^{\mathrm{G}}\rightharpoonup V^{\mathrm{T}}$ greedily by selecting the highest-scoring pairs without reuse of $g$ or $t$ (pairs with zero scores are ignored). We call it a partial injective or one-to-one mapping because there may be extra elements in $V^G$ which do not match any element in $V^T$ or vice-versa.
Then, we replace whole-word occurrences of each $g$ by $\mu(g)$ in \texttt{generated\_code}, yielding \texttt{aligned\_generated\_code}. We sort longer names first. For example, let's assume the mapping $x\rightarrow \text{demand}$, and $x^{'} \rightarrow \text{demand\_i}$. If we replace $x$ first, then $x^{'}\rightarrow \text{demand}^{'}$, which is wrong. 

\textit{Category-wise matching:}
we extract components from \texttt{aligned\_generated\_code}:
$
\widehat V^{\mathrm{G}}, \widehat C^{\mathrm{G}},$ and $ \widehat O^{\mathrm{G}}.
$
Counts are defined as follows:
\begin{itemize}
\item Variables: $M_{\mathrm{var}}=\bigl|\,V^{\mathrm{T}}\cap \widehat V^{\mathrm{G}}\,\bigr|$.
\item Constraints: a true constraint $c^{\mathrm{T}}\!\in\!C^{\mathrm{T}}$ is matched if there exists $\hat c\!\in\!\widehat C^{\mathrm{G}}$ such that either
$$
\mathcal{N}_{\mathrm{idx}}(c^{\mathrm{T}})=\mathcal{N}_{\mathrm{idx}}(\hat c),
$$
or, for some $(g,t)\in\mathrm{graph}(\mu)$,
$$
\mathcal{N}_{\mathrm{var}}(\mathcal{N}_{\mathrm{idx}}(c^{\mathrm{T}});t)
=
\mathcal{N}_{\mathrm{var}}(\mathcal{N}_{\mathrm{idx}}(\hat c);g),
\quad\text{or}\quad
\mathcal{R}(\cdot)\ \text{of the two sides simplify to equal forms}.
$$
Let $M_{\mathrm{cons}}$ be the number of matched constraints.
\item Objective: treat $O^{\mathrm{T}}$ and $\widehat O^{\mathrm{G}}$ as singleton sets and apply the same logic to obtain $M_{\mathrm{obj}}\in\{0,1\}$.
\end{itemize}
Let $T_{\mathrm{var}}=|V^{\mathrm{T}}|$, $T_{\mathrm{cons}}=|C^{\mathrm{T}}|$, $T_{\mathrm{obj}}=|O^{\mathrm{T}}|$. The percentages are
$$
P_{\mathrm{var}}=100\cdot \frac{M_{\mathrm{var}}}{T_{\mathrm{var}}},\qquad
P_{\mathrm{cons}}=100\cdot \frac{M_{\mathrm{cons}}}{T_{\mathrm{cons}}},\qquad
P_{\mathrm{obj}}=100\cdot \frac{M_{\mathrm{obj}}}{T_{\mathrm{obj}}}.
$$
The total matched percentage is
$
P_{\mathrm{total}}=100\cdot \frac{M_{\mathrm{var}}+M_{\mathrm{cons}}+M_{\mathrm{obj}}}{T_{\mathrm{var}}+T_{\mathrm{cons}}+T_{\mathrm{obj}}}.
$
Finally, let
$
G=|\widehat V^{\mathrm{G}}|+|\widehat C^{\mathrm{G}}|+|\widehat O^{\mathrm{G}}|
$
be the number of generated components, and let $E$ be the number of generated components that did not participate in any match; we report
$
P_{\mathrm{extra}}=100\cdot \frac{E}{G}.
$

\section{Solution and analysis}\label{3}

In this section, we analyze the solutions obtained from four recent ChatGPT models: GPT-3.5-Turbo, GPT-3.5-16K, GPT-4, and GPT-4-Turbo, using the prompting methods listed in Table \ref{tab:prompting_methods}. The input problems are summarized in Table \ref{tab:problem-summary}. 
We use zero temperature\footnote{Temperature is a hyperparameter with a range in [0,2], influencing the randomness of the model. Lower temperatures result in more deterministic outputs, and higher temperatures increase randomness \cite{openai-chat-completions-api}.} hyperparameter throughout the experiments to limit superficial variability that would otherwise confound structural comparison and soft scoring, following best practices in recent LLM benchmarks (e.g., \cite{lee2024optimize, xiao2023chain}). We conducted 360 runs, i.e., a combination of nine problems, two instances for each problem, four ChatGPT models, and five prompting methods, and 200 runs for analyzing the reproducibility of results for electricity planning problem instances, i.e., a combination of two problem instances, four GPT models, and five prompting methods, and 10 runs for each combination.

Parts of the implementation code build upon open-source code from the publication \cite{xiao2023chain}. 
 In interpreting the results, we compare model and prompt performance only relative to one another. All statements are therefore strictly relative to the specific models and prompting methods evaluated in this paper, not absolute claims about broader performance.

\begin{table}[H]
\centering
\caption{Overview of stochastic problem used in the experiments.}
\label{tab:problem-summary}
\begin{tabular}{p{4.5cm} p{2cm} p{5cm} p{2cm} p{1cm}}
\hline
Problem Name & Instances &Category & Adapted &Taken \\
\hline
Electricity planning &2 & SMILP-2 & & \cite{birge_louveaux_datasets} \\
Biweekly fleet size planning &2 & SMILP-2 & & \cite{birge_louveaux_datasets}\\
Weekly fleet size planning & 2& SMILP-2 & & \cite{birge_louveaux_datasets}\\
Equipment leasing & 2& DMILP-2 &\cite{birge_louveaux_datasets} & \\
Production planning &2 & DMILP-2 &\cite{birge_louveaux_datasets} & \\
Emergency supply &2 & Joint chance-constrained & \cite{klein2020chance}& \\
Production demand &2 & Joint chance-constrained & \cite{klein2020chance}& \\
Cold storage &2 & Individual chance-constrained & \cite{shapiro2007tutorial}& \\
Supply allocation & 2& Individual chance-constrained &\cite{qiu2014covering} & \\
\hline
\end{tabular}
\end{table}

\begin{figure}[H]
\centering
\includegraphics[width=1\textwidth]{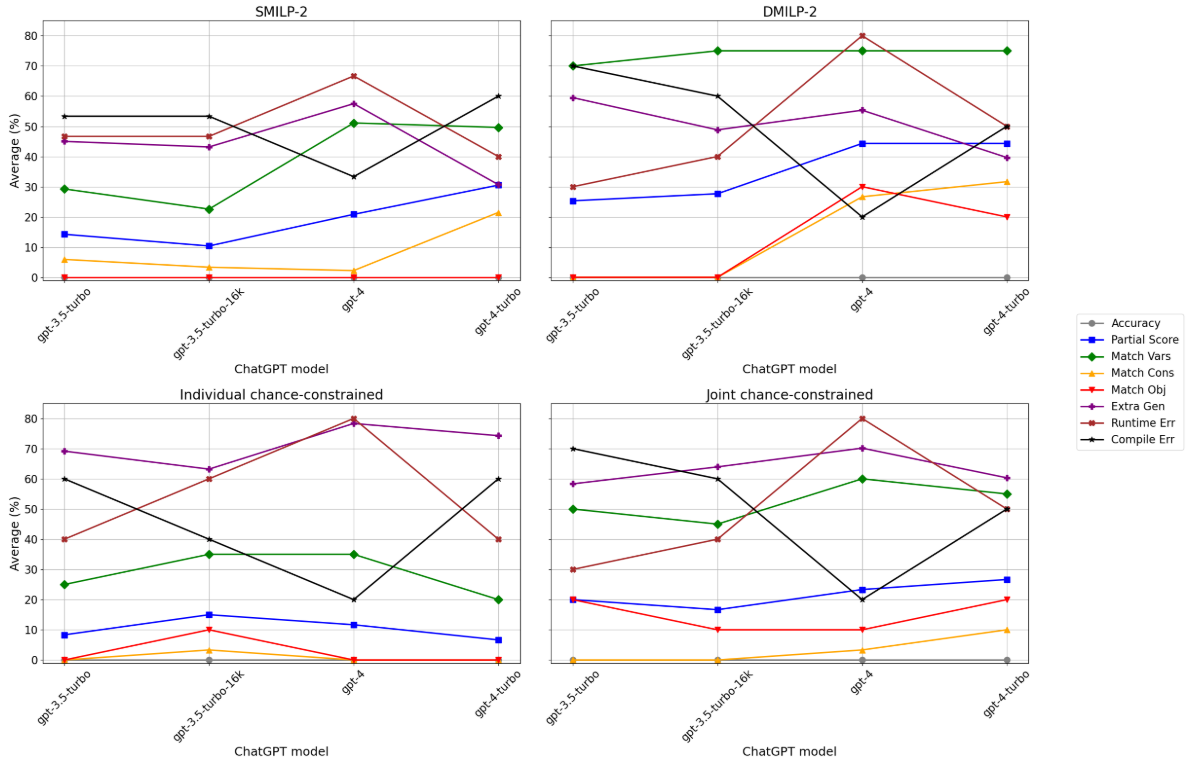}
\caption{Average performance metrics of ChatGPT models across problem categories.}
\label{fig:metricsbymodelbycategory}
\end{figure}

Figure~\ref{fig:metricsbymodelbycategory} summarizes the average performance (across instances and problem categories) of the four ChatGPT models across the problem categories.

The accuracy metric remained zero for all problem instances and categories across all prompting and GPT models. 

On average, GPT-4-Turbo achieved the best performance for SMILP-2 problems among GPT models with the highest partial score, matching scores for variables and constraints, the least extra element generation, and runtime error scores. Objective matching score remained zero for all GPT models in this problem category. Three other GPT models showed fluctuating results in terms of matching scores, but GPT-4 achieved the second-best scores in terms of variable matching and partial scores. 

The GPT-3.5-Turbo-16k performs better than the other models on individual chance-constrained problems in terms of partial score, variable, and constraints matching scores. However, the partial score remains below $20\%$ for this problem category across all GPT models.

 On average, GPT-4-Turbo achieved the best performance for DMILP-2 problems among GPT models with the highest partial score, matching scores for variables and constraints, and the least extra element generation score. GPT-4 performed almost as well as GPT-4-Turbo, with a lower constraint matching score and a higher objective matching score. 
 
 Again, on average, GPT-4-Turbo outperformed other models on joint chance constrained models in terms of partial score, objective, and constraints matching scores. GPT-4 scored highest in variable matching. 
 
Across categories, the lowest extra element generation rates vary by model. For DMILP-2, the lowest rate is achieved by GPT-4-Turbo. In the individual chance-constrained category, GPT-3.5-Turbo-16k performs best. The joint chance-constrained category shows the lowest extra element generation levels with GPT-3.5-Turbo. For SMILP-2, GPT-4-Turbo offers the most reliable behavior.

\begin{figure}[H]
\centering
\includegraphics[width=1\textwidth]{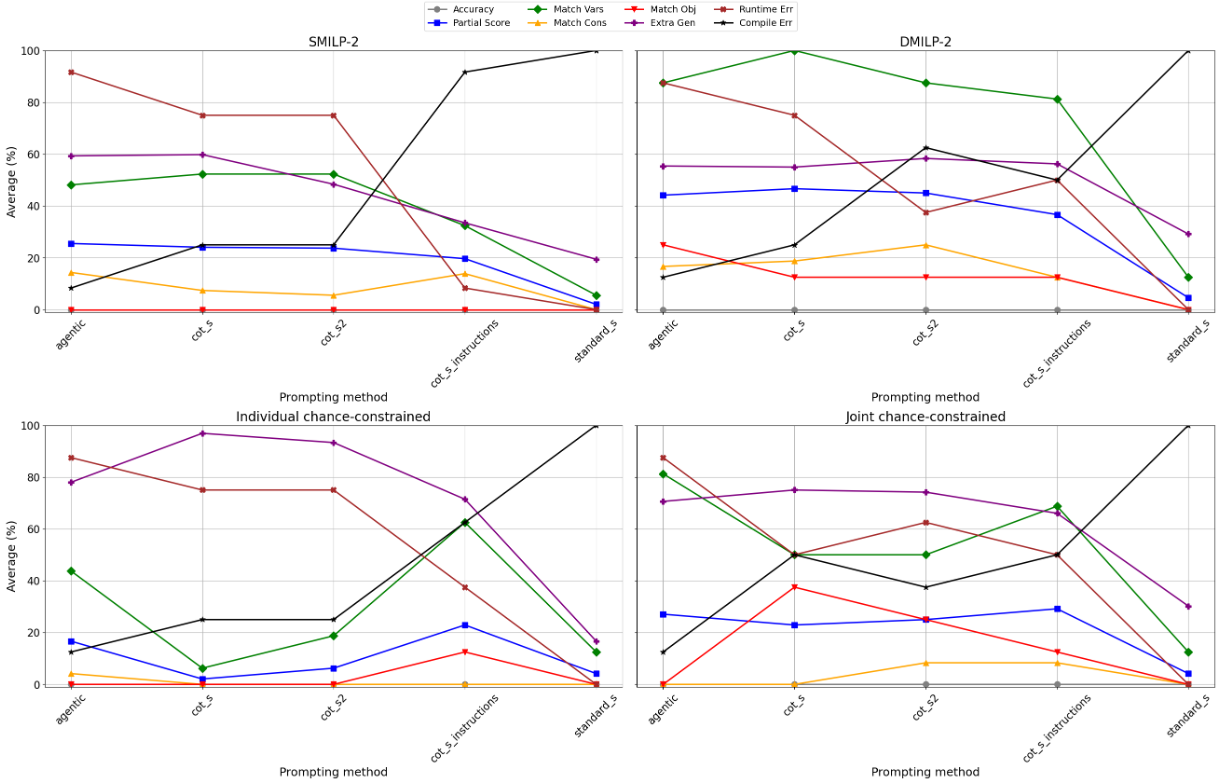}
\caption{Average performance metrics of prompting methods across problem categories.}
 \label{fig:performancepromptingproblem}
\end{figure}

According to Figure~\ref{fig:performancepromptingproblem}, the prompting methods exhibit clear differences in modeling quality. 
On average, \texttt{agentic} outperforms other methods in terms of partial score and constraint matching scores, with the highest extra element generation score (negative effect) on the SMILP-2 problem category.  On the DMILP-2 problem category, \texttt{cot\_s} achieves the highest partial and variable matching scores. Highest matching scores for objective and constraints obtained by \texttt{agentic} and \texttt{cot\_s2}, respectively.

As for individual chance-constrained problems, \texttt{cot\_instructions} outperforms among models with the highest partial, variable, and objective matching scores, while \texttt{agentic}, as second second-best model, achieves the highest constraint matching score. Similarly, on joint chance-constrained models, these two prompting methods achieved the highest partial scores and variable matching score. In particular, \texttt{cot\_s2} outperformed \texttt{agentic} in constraint matching score, and \texttt{cot\_s} achieved the highest objective matching score among others. 

Finally, the \texttt{standard\_s} baseline performs substantially worse than all other methods, with low matching scores throughout. In many runs, this prompt did not return a complete Python code, which resulted in low extra-element generation rates.

\begin{figure}[H]
\centering
\includegraphics[width=0.9\textwidth]{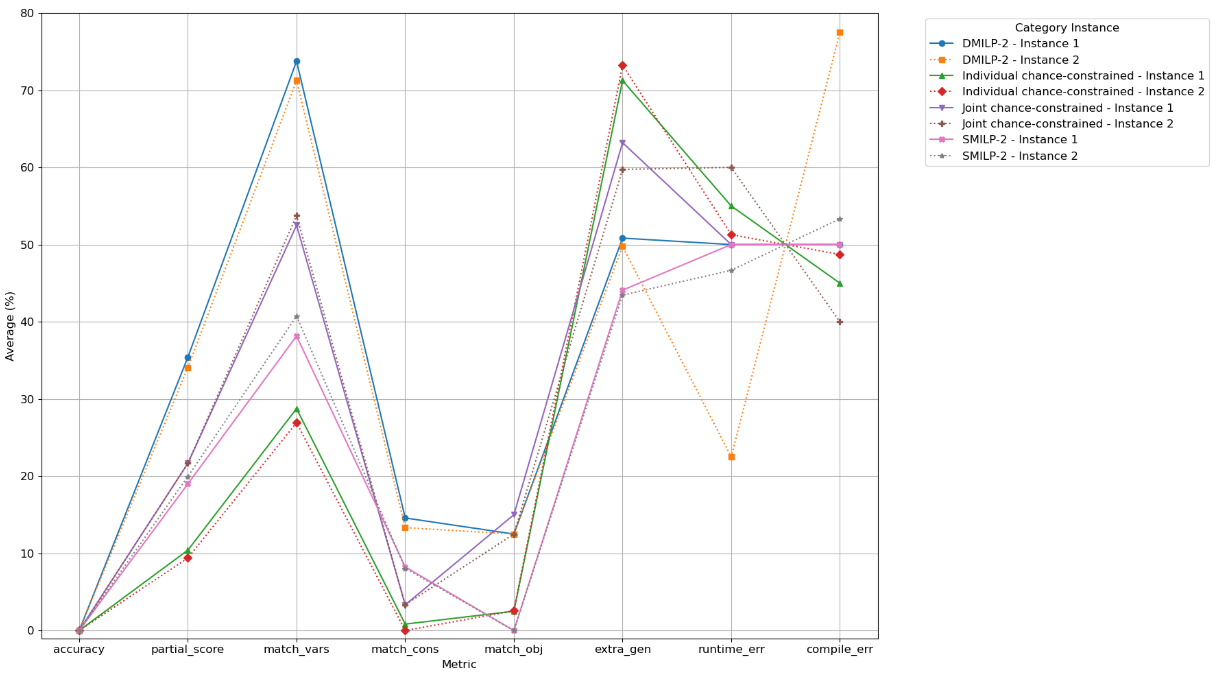}
\caption{Average metrics by problem category instance.}
\label{fig:instance}
\end{figure}

Figure~\ref{fig:instance} reports instance-level performance across the four problem categories.
The variation in results is less than $5\%$ in all problem instances and metrics except for runtime error and compile error. Runtime and compile errors' variation is most pronounced for DMILP-2 and joint chance-constrained instances. 

\begin{figure}[H]
\centering
   \includegraphics[width=1\textwidth]{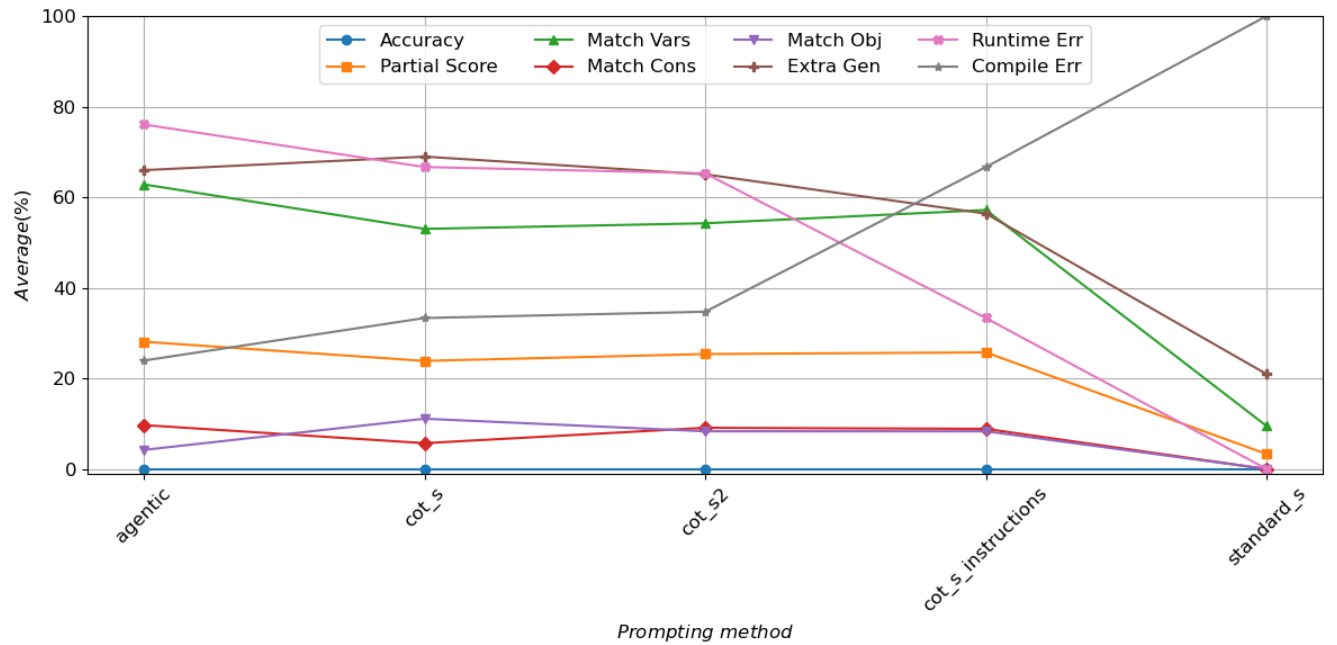}
\caption{Comparison of average performance metrics across different prompting methods.}
  \label{fig:prompting}
\end{figure}

Figure~\ref{fig:prompting} reports the average performance of the prompting methods on all problem instances and across all metrics. 
The \texttt{agentic} prompting method achieves the highest partial and matching scores across all problem instances on average. It achieves the second-lowest extra-element generation rate among prompting methods. It is to be mentioned that we do not consider the extra element generation score of  \texttt{standard\_s}, as in most runs, it did not generate model code output, thereby having the smallest score among prompting methods. The second-best prompting method is the \texttt{cot\_instructions} with the lowest extra element generation score.

The \texttt{cot\_s\_instructions} method achieves the highest variable-matching score, though it also produces a relatively high amount of extra elements. The \texttt{cot\_s2} method shows a balanced pattern, with steady partial scores and moderate objective matching. The \texttt{agentic} approach performs reasonably well, with higher partial scores and a lower compilation error rate compared to the other structured prompts.
\texttt{cot\_s2} and \texttt{cot\_s} performed similarly with \texttt{cot\_s} being better in objective matching score and \texttt{cot\_s2} outperforming in constraint matching score.

In contrast, \texttt{standard\_s} has the weakest performance. It obtains the lowest matching scores and partial scores. 
Overall, complex prompting methods provided better outputs in comparison to simpler prompting methods.

\begin{figure}[H]
  \centering
   \includegraphics[width=1\textwidth]{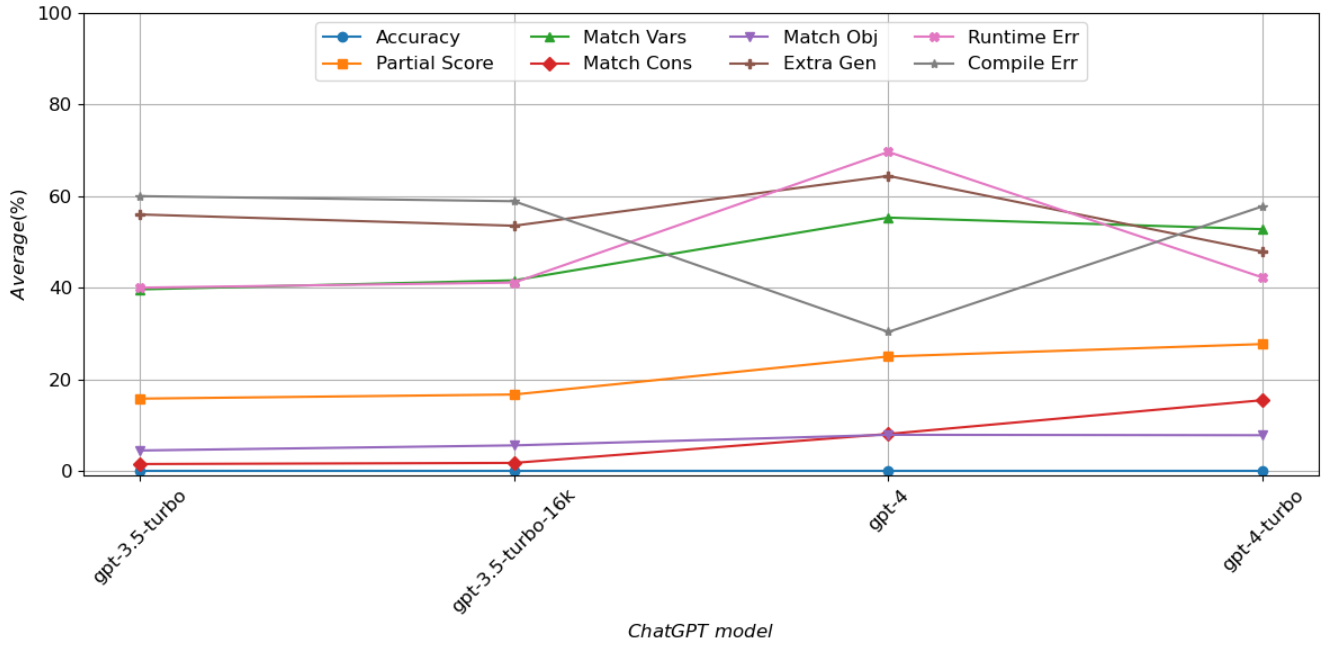}
  \caption{Average metrics per GPT model.}
  \label{fig:chat}
\end{figure}

Figure~\ref{fig:chat} shows the average performance of GPT models on all problem instances and across all metrics. The GPT-4-Turbo model performs the best overall, except for the average variable matching score which GPT-4 is the best. It reaches the highest partial score and the highest objective matching. The GPT-4 model performs moderately well, though it produces more extra elements than GPT-4-Turbo.

The two GPT-3.5 variants perform weaker on most metrics. They show lower partial and matching scores and generate more extra elements compared to the GPT-4 and GPT-4-Turbo models. However, they show similar scores on extra element generation.

\begin{figure}[H]
    \centering
   \includegraphics[width=1\textwidth]{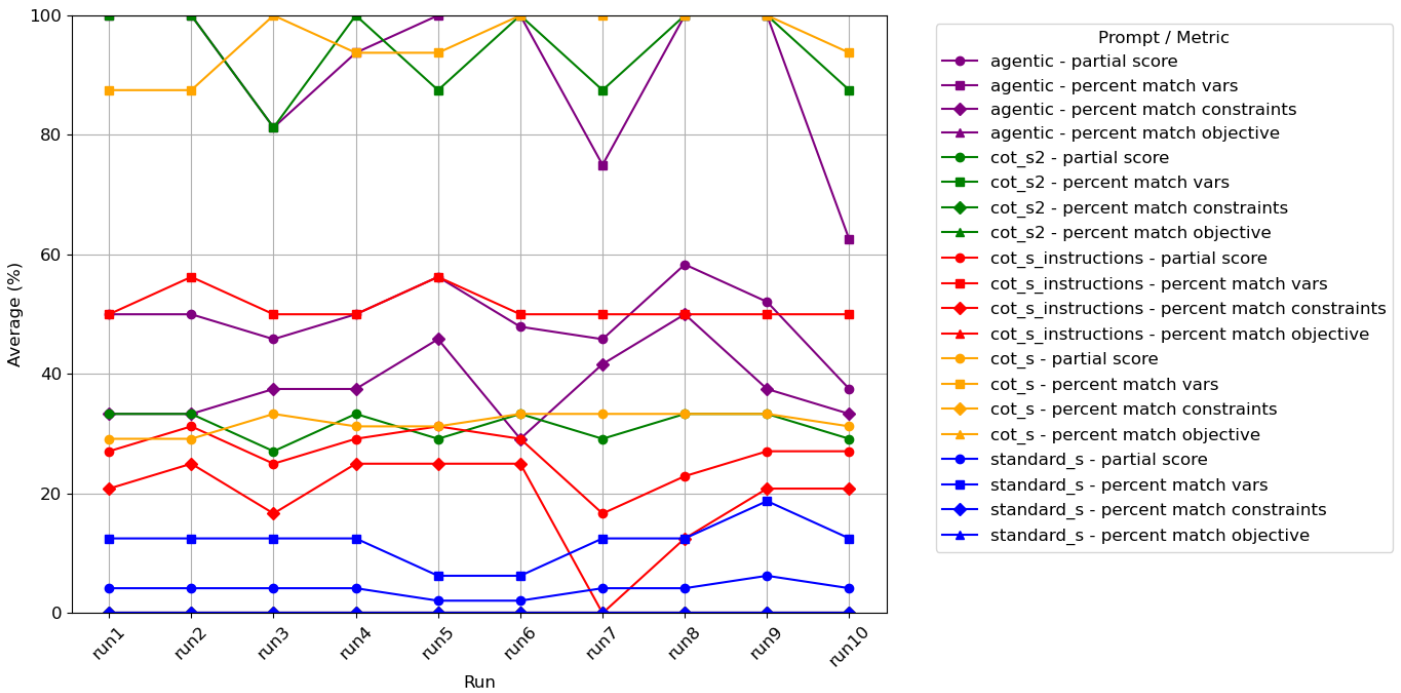}
    \caption{Prompting method performance trends across runs (averaged over instances).}
     \label{fig:run1}
\end{figure}

Figure~\ref{fig:run1} shows the consistency of the prompting methods across ten runs, averaged over the SMILP-2 instances of the electricity planning problem. Variable matching and partial scores remain relatively stable over runs, with \texttt{cot\_s\_instructions} and \texttt{cot\_s} showing the most consistent behavior with less than $10\%$ variation, while \texttt{agentic} demonstrates variations above $20\%$ in runs seven and 10.

Objective matching stays zero in all runs for all prompting methods. Constraint matching is highly unstable and often variations pass $20\%$ in a number of runs for \texttt{agentic} and \texttt{cot\_instructions}.

\begin{figure}[H]
\centering
   \includegraphics[width=0.85\textwidth]{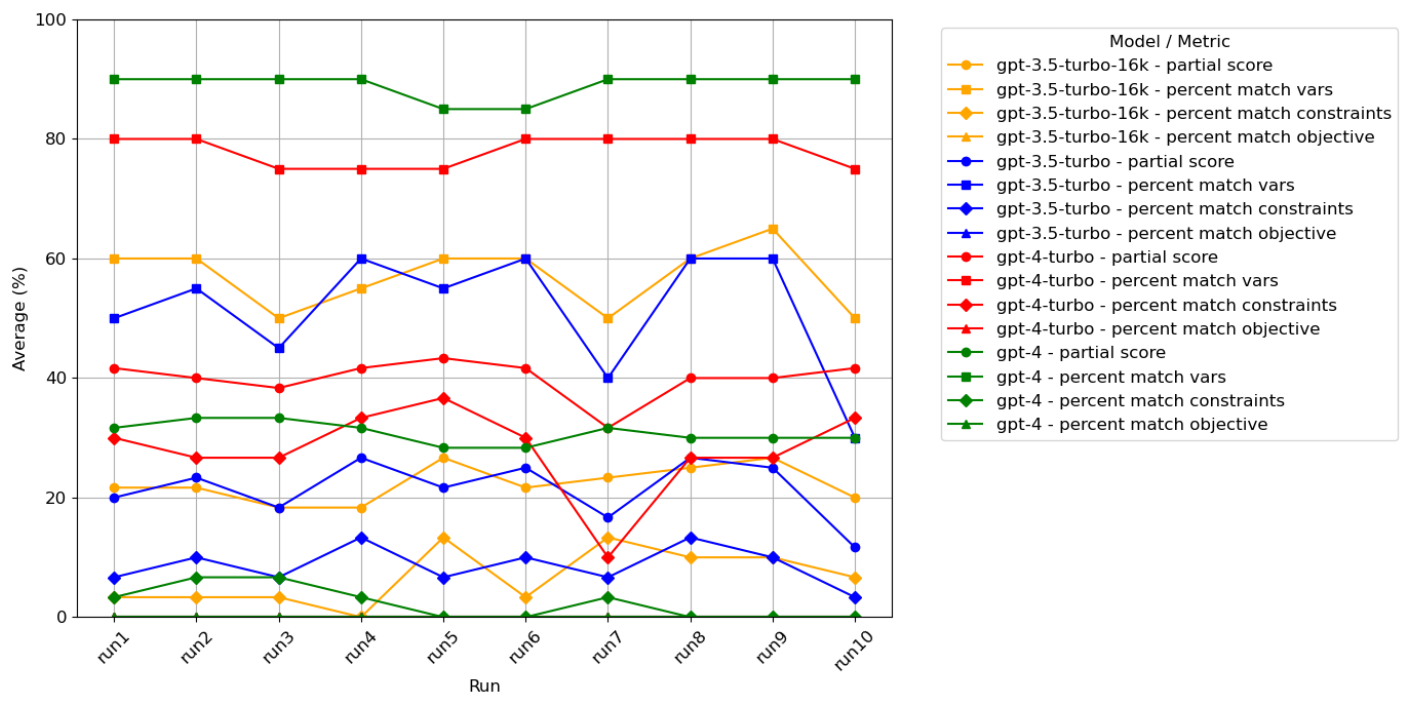}
\caption{Model-wise performance trends across runs (averaged over instances).}
\label{fig:fig8}
\end{figure}

Figure~\ref{fig:fig8} shows the stability of the GPT models across ten runs, averaged over the electricity planning SMILP-2 instances. GPT-4-Turbo and GPT-4 remain stable in partial score and variable matching scores with only small variations across runs. However, they show high variations in constraint matching scores.

GPT-3.5 and GPT-3.5-Turbo-16k maintain a consistent partial score and constraint matching score, but show noticeable variation in variable matching score.

\begin{figure}[H]
  \centering
  \includegraphics[ scale = 0.83]{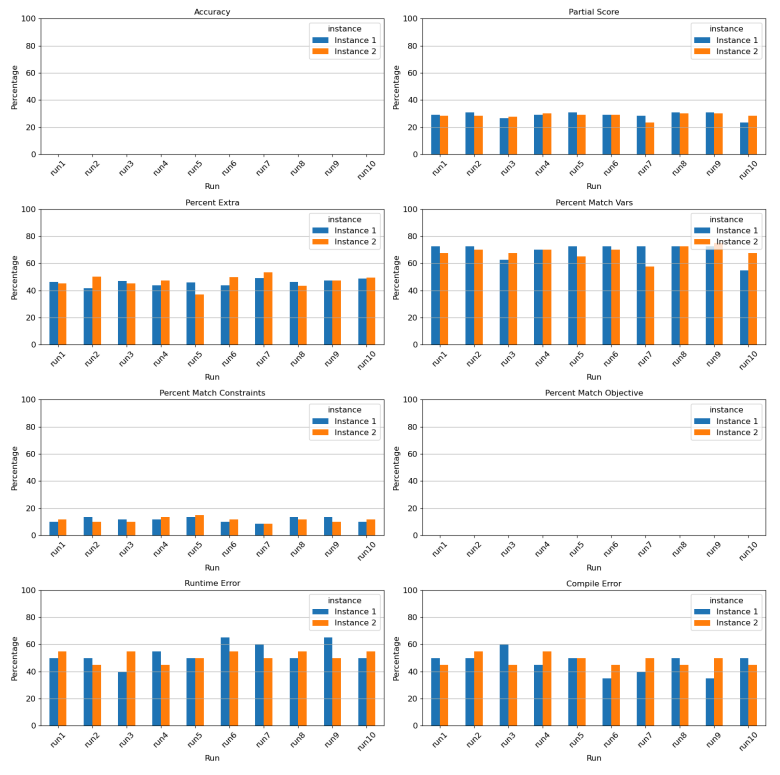}
  \caption{Metric reproducibility across runs for two instances of the SMILP-2 problem.}
  \label{fig:repro}
\end{figure}

Figure \ref{fig:repro} reports the reproducibility of several metrics across ten runs for two instances of the electricity planning SMILP-2 problem. Partial score remains relatively stable for both instances, with only fluctuations of less than $10\%$ across runs. Variable matching is also consistent with variations less than $10\%$ except for runs seven and 10.
Extra element generation shows moderate variation but follows similar patterns across the two instances.
Constraint matching remains below $20\%$ in all runs for both instances. Objective matching and accuracy remained zero for both instances in all runs.
The extra element generation score also remained consistent across runs for both instances, with higher fluctuations for the second instance, especially in run five.
The error metrics display larger discrepancies between instances. Both instances exhibit high runtime and compile error rates depending on the runs.

In summary, the results show that partial score, variable, and constraint matching scores are comparatively reproducible across runs.

\subsection{Discussion on the results}

\subsubsection{Limitations}

While LLMs show the ability to translate verbal problem descriptions into formal SO formulations, they face notable limitations when applied to complex instances. We were unable to generate fully correct and executable code formulations. Despite achieving a range of structural and algebraic similarity scores, many outputs contained syntax errors, incomplete solver definitions, or invalid variable declarations that led to compile or runtime errors (e.g., Figure~\ref{fig:instance}).

Multi-stage stochastic programs (more than two stages) remain beyond the current scoring system scope. Furthermore, correct formulations with different algebraic formulations can still hold true for an optimization problem, in which case, matching scores will be underrepresented. Additionally, throughout our experiments, we observed that indices, coefficients, and variables often employ text-intensive indexing and naming, which requires more sensitive parsing of model elements to avoid underestimation of scores.

Third, reproducibility of results remains an ongoing challenge. Due to Figure \ref{fig:repro}, different runs of the same instance or different instances of a problem can show low or high fidelity depending on components of the problem.

% \subsubsection{Durability and benefits of the proposed metric}

% A key contribution of this paper is the soft scoring metric, which offers a durable and interpretable way to evaluate LLM-generated optimization models. Unlike binary checks, it measures partial correctness through the matching of variables, objectives, and constraints. The metric tolerates differences in notation and structure, making it valid across prompt types, LLM versions, and stochastic model classes. It requires no retraining or manual adjustment. This durability allows consistent benchmarking of model quality, even when formulations differ in form. 

% \smallskip
% \noindent
% \subsubsection{Prompting and framework performance}
% This paper demonstrates that with carefully engineered prompts, particularly the \texttt{cot\_s\_instructions} variant, and structured workflows such as our multi-agent framework, LLMs like GPT-4-Turbo can extract, formulate, and implement SMILP-2 models from NL. We observed reasonable variable and objective matching scores for SMILP-2 problems, especially under the \texttt{cot\_s\_instructions} and agentic prompting methods. 

\subsubsection{Real-world applicability of our method and an example of an SMILP-2 supply chain model}

Our framework enables real-world industries to model and evaluate SO problems using NL descriptions alone rapidly. This is particularly valuable in sectors such as supply chains and transportation, where decisions must be made under uncertainty and under time pressure. 
Our soft scoring metric allows practitioners to quantitatively evaluate LLM-generated implementation code even when exact syntax or notation varies to some extent. It measures structural correctness, variable and objective matching, and highlights errors like extra elements or constraint mismatches. This enables transparent model validation. Additionally, it enables the identification and fixation of the errors in the generated code quickly by spotting the deficient parts of the code.

We take an SMILP-2 model with adaptations from \cite{santoso2005stochastic} for a supply chain network. \cite{santoso2005stochastic} implemented the model for a company shipping cardboard packages to breweries and soft-drink producers, adapted from \cite{dogan1999primal}. Similar problems to this case can be found in recent literature (e.g.,  \cite{alinezhad2022fuzzy, koca2023two}). 

Let $P$ be the set of potential facilities (e.g., plants, finishers, or warehouses) where in- and outflow are allowed for each facility in this set, $S\subset P$ be the set of suppliers, $C$ be the set of customer nodes. The  
$A\subseteq (P\times P)\cup (P\times C)$ is the set of feasible arcs such that shipments within facilities, and from facilities to customers are allowed, $K$, is the product set, and $\Omega$ is a finite scenario set
with probability $p_\omega$ for scenario $\omega \in \Omega$.

Fixed opening costs for facilities are $c_i$, $i \in P$.
Scenario $\omega$ provides transportation and processing costs
$q_{ijk}^\omega$ for product $k\in K$ shipped on arc $(i,j)$, supply limits $s_{ik}^\omega$ of product $k$ at supplier $i\in S$, demand of product $d_{jk}^\omega$ and shortage penalties $h_{jk}^\omega$ of product $k$ at customer $j\in C$.

Binary variables $y_i$ indicate which facilities are opened, i.e., $y_i = 1$, if facility $i$ is opened and 
0 otherwise.
For each scenario $\omega \in \Omega$, the model chooses flows
$x_{ijk}^\omega$, i.e., flow of product $k$ on arc $(i,j)$ under scenario $\omega$, and shortage quantities
$ z_{jk}^\omega$, i.e., shortage of product $k$ at customer $j$ under scenario $\omega$. $u_{ik}^{\omega}$ is the injected supply of product $k$ by supplier $i$ to the supply network at scenario $\omega$. 
The compact SMILP-2 formulation is
\begin{align*}
  \min_{y}
  \quad
  & \sum_{i \in P} c_i y_i
    +
    \mathbb{E} \big[ Q(y,\omega) \big]
     \\
  \text{s.t.} \quad
  & y_i \in \{0,1\},\ \forall i \in P,
\end{align*}
where the recourse function $Q(y,\omega)$ is
\begin{align*}
  Q(y,\omega)
  :=
  \min_{x^\omega,\,z^\omega, u^\omega}
  \quad
  &
  \sum_{k\in K} \sum_{(i,j) \in A} q_{ijk}^\omega x_{ijk}^\omega
  + \sum_{j\in C, k\in K} h_{jk}^\omega z_{jk}^\omega
  \label{eq:ts-recourse-obj}
  \\
  \text{s.t.} \quad
   &
  \sum_{i:(i,j)\in A} x_{ijk}^\omega- \sum_{l:(j,l)\in A} x_{jlk}^\omega
  = 0,
   \forall j \in P\backslash S,\ \forall k\in K
  \text{(flow conservation at non-suppliers),} \\
  &
  \sum_{i:(i,j)\in A} x_{ijk}^\omega +
  u_{jk}^{\omega}- \sum_{l:(j,l)\in A} x_{jlk}^\omega
  = 0,
   \forall j \in S,\ \forall k\in K
  \text{(flow conservation at suppliers),} \\
  &u_{ik}^{\omega}\leq s_{ik}^{\omega} y_i, \forall i\in S, k\in K (\text{injected supply availability}),           \\
  &
  \sum_{i: (i,j)\in A} x_{ijk}^\omega + z_{jk}^\omega
  \ge d_{jk}^\omega, \forall j \in C,\ \forall k\in K ( \text{demand satisfaction with allowed shortage}), \\
  &
  \sum_{j: (i,j)\in A} x_{ijk}^\omega
  \le M\cdot y_i,
   \forall i \in P,\ \forall k\in K, (\text{no outflow for un-chosen supplier}), \\
    &
  \sum_{i: (i,j)\in A} x_{ijk}^\omega
  \le M\cdot y_j,
   \forall j \in P,\ \forall k\in K, (\text{no inflow for un-chosen supplier}), \\
  &
  x_{ijk}^\omega \ge 0, \forall (i,j)\in A,\qquad
  z_{jk}^\omega \ge 0, \forall j\in C,k\in K\qquad u_{ik}^{\omega}\geq 0, \forall i\in P, k\in K.
\end{align*}

First-stage decisions select facilities.
Second-stage decisions adapt flows and shortages to uncertainty in demand, supply availability, and transportation costs.
Here is the problem description we ask from GPT-4-Turbo using \texttt{agentic} prompting method: 

\begin{tcolorbox}[promptbox, title={Problem description: supply chain SMILP-2 problem, adapted from \cite{santoso2005stochastic}}]
Consider the following two-stage stochastic mixed-integer linear programming (SMILP-2) for the design of a supply chain network under uncertainty: 

The supply chain consists of a set of potential facilities $P$, a subset $S \subset P$ of supplier facilities, a set of customer nodes $C$, and a set of feasible arcs $A \subseteq (P \times P) \cup (P \times C)$ through which shipments within facilities and from facilities to customers may flow. The system handles a set of products $K$, and uncertainty is represented through a finite scenario set $\Omega$, where each scenario $\omega \in \Omega$ occurs with probability $p_\omega$.

Each potential facility $i \in P$ has a fixed opening cost $c_i$. In scenario $\omega$, the uncertain data include transportation and processing costs $q_{ijk}^\omega$ for shipping product $k$ through arc $(i,j)\in A$, supply availability $s_{ik}^\omega$ at supplier $i \in S$, customer demand $d_{jk}^\omega$ at $j \in C$, and shortage penalties $h_{jk}^\omega$ for unsatisfied demand at customer $j \in C$.

First-stage decision variables $y_i$ indicate whether facility $i$ is opened. After scenario $\omega$ is realized, second-stage decisions include shipment flows $x_{ijk}^\omega$, injected supplies $u_{ik}^\omega$ at supplier nodes, and shortage amount $z_{jk}^\omega$ at customers. These second-stage decisions must satisfy flow conservation at all facilities, supplier supply availability limits, demand satisfaction with allowed shortages, and linking constraints that prevent any inflow or outflow through facilities that are not opened.

The objective is to minimize the sum of fixed facility opening costs and the expected second-stage operational cost, which consists of transportation and processing costs and penalties for unmet customer demand.

\end{tcolorbox}

We include the generated model's code in Listing~\ref{lst:llm-generated-model}.
 We conducted 15 runs and selected the run with the highest partial score.  
 
Table~\ref{tab:llm-aggregate-metrics} summarizes its evaluation results.

\begin{table}[H]
    \centering
    \caption{Evaluation metrics for GPT-4-Turbo generated implementation code.}
    \label{tab:llm-aggregate-metrics}
    \begin{tabular}{lr}
        \hline
        Metric & Value (\%) \\
        \hline
        Percent extra components in generated model & 36.36 \\
        Percent match in variables                 & 100.00 \\
        Percent match in constraints               & 0.00 \\
        Percent match in objective                 & 0.00 \\
        Compile error rate                         & 100.00 \\
        Runtime error rate                         & 0.00 \\
        \hline
    \end{tabular}
\end{table}

The evaluation shows that the LLM correctly identified all variables and index sets. This is useful because variable and index set extraction are often a time-consuming part for non-experts.

The main issue appears in the objective function and constraint set. Putting indexing of variables and sum loops in the objective function aside, the objective function correctly interprets elements of expected costs to be minimized. However, the inconsistency in indexing variables and summing over the wrong indices and loops resulted in a zero objective matching score, although index sets and variable indices have been defined correctly in the first place. The performance of the metric in the evaluation of these issues has been shown in \ref{lst:llm-generated-model-evaluation}.

As for constraints, structurally or algebraically, outflow and inflow constraints are wrong. This is again due to the wrong indexing of variables and the misinterpretation of outflow and inflow logic.  
Regarding flow conservation at suppliers, again generated code uses inconsistent feasible arc syntax and references the wrong variable keys. However, if the metric were index-agnostic, i.e., masking indices completely, this constraint set would match the true constraint set, but in this version, it is very sensitive to such inconsistencies. This applies to demand satisfaction constraints as well, where wrong scenario data indexing has happened in the generated code. For supply limits constraints, the generated code misses supplier capacity limit parameters in addition to wrong indexings. Finally, inflow and outflow gating constraints for chosen and unchosen facilities have been conceptually modeled differently from the true code. Although indexing is wrong for this constraint set in the generated code, the metric will not be able to capture algebraic equivalence for this constraint set.

For the industry, this means LLMs can help produce an initial model quickly, but the model logic still needs expert review. The benefit is speed and accessibility; the limitation is reliability in structurally complex parts of the formulation.  A helpful reference on how to model SO models applied in industry is \cite{king2012modeling}. LLMs can be trained and fine-tuned on such resources to effectively model SO problems.

\begin{lstlisting}[style=gurobicode, numbers=none,
caption={GPT-4-Turbo generated Gurobi model for the SMILP-2 supply chain instance.},
label={lst:llm-generated-model}
]
from gurobipy import Model, GRB, quicksum

def Industrial_example(P, S, C, K, A, c, scenarios, M):
    model = Model("SMILP2_Extensive")
    model.setParam("OutputFlag", 0)
    model.setParam("MIPGap", 0.01)  # Setting a MIP gap for early stopping if the gap is small enough

    # Sets and indices
    P = list(P)  # Potential facilities
    S = list(S)  # Supplier facilities
    C = list(C)  # Customer nodes
    K = list(K)  # Products
    A = [tuple(e) for e in A]  # Feasible arcs
    W = range(len(scenarios))  # Scenarios

    # Variables
    y = model.addVars(P, vtype=GRB.BINARY, name="y")  # Facility opening decisions
    x = model.addVars(A, K, W, vtype=GRB.CONTINUOUS, name="x")  # Shipment flows
    u = model.addVars(S, K, W, vtype=GRB.CONTINUOUS, name="u")  # Injected supplies
    z = model.addVars(C, K, W, vtype=GRB.CONTINUOUS, name="z")  # Shortage amounts

    # Objective function
    model.setObjective(
        quicksum(c[i] * y[i] for i in P) +
        quicksum(scenarios[w]['probability'] * (
            quicksum(scenarios[w]['q'][i, j, k] * x[i, j, k, w] for i, j, k in A if (i, j) in A) +
            quicksum(scenarios[w]['h'][j, k] * z[j, k, w] for j in C for k in K)
        ) for w in W),
        GRB.MINIMIZE
    )

    # Constraints
    # Flow conservation at non-supplier facilities
    model.addConstrs(
        (quicksum(x[i, j, k, w] for i, j in A if j == p) == quicksum(x[j, i, k, w] for j, i in A if j == p) 
         for p in P if p not in S for k in K for w in W),
        name="flow_conservation_non_suppliers"
    )

    # Flow conservation at supplier facilities
    model.addConstrs(
        (quicksum(x[i, j, k, w] for i, j in A if j == s) + u[s, k, w] == quicksum(x[j, i, k, w] for j, i in A if j == s) 
         for s in S for k in K for w in W),
        name="flow_conservation_suppliers"
    )

    # Demand satisfaction
    model.addConstrs(
        (quicksum(x[i, j, k, w] for i, j in A if j == c) + z[c, k, w] >= scenarios[w]['d'][c, k] 
         for c in C for k in K for w in W),
        name="demand_satisfaction"
    )

    # Supply limits
    model.addConstrs(
        (u[s, k, w] <= scenarios[w]['s'][s, k] * y[s] for s in S for k in K for w in W),
        name="supply_limits"
    )

    # Linking constraints (both from and to facilities)
    model.addConstrs(
        (x[i, j, k, w] <= M * y[i] for i, j, k in A for w in W),
        name="linking_constraints_from"
    )
    model.addConstrs(
        (x[j, i, k, w] <= M * y[i] for i, j, k in A for w in W),
        name="linking_constraints_to"
    )

    # Optimize the model
    try:
        model.optimize()
        if model.Status == GRB.OPTIMAL:
            return model.ObjVal
        else:
            print("Optimization was stopped with status:", model.Status)
    except Exception as e:
        print("An error occurred during optimization:", str(e))

    return None

# Example usage
P = ['Facility1', 'Facility2']
S = ['Facility1']
C = ['Customer1', 'Customer2']
K = ['Product1', 'Product2']
A = [('Facility1', 'Customer1'), ('Facility1', 'Customer2'), ('Facility2', 'Customer1')]
c = {'Facility1': 1000, 'Facility2': 1500}
scenarios = [
    {'probability': 0.5, 'q': {('Facility1', 'Customer1', 'Product1'): 2, ('Facility1', 'Customer2', 'Product2'): 3},
     'h': {'Customer1': {'Product1': 10}, 'Customer2': {'Product2': 15}},
     'd': {'Customer1': {'Product1': 30}, 'Customer2': {'Product2': 20}},
     's': {'Facility1': {'Product1': 40, 'Product2': 50}}
    }
]
M = 10000

optimal_cost = Industrial_example(P, S, C, K, A, c, scenarios, M)
print("Optimal expected total cost:", optimal_cost)
\end{lstlisting}

\begin{lstlisting}[style=gurobicode, numbers=none,
caption={GPT-4-Turbo generated Gurobi model for the SMILP-2 supply chain instance evaluation.},
label={lst:llm-generated-model-evaluation}
]
Mapping (generated -> true): {'x': 'x', 'y': 'y', 'z': 'z', 'u': 'u'}
Matched variables:   4/4 -> ['u', 'x', 'y', 'z']
Extra generated vars (0): []
Matched constraints: 0/6
Constraint matches (true -> gen): (none)
TRUE constraints (6):
  [T1] UNMATCHED: quicksum(x[a, k, w] for a in inP[p]) + u[S_index[p], k, w] - quicksum(x[a, k, w] for a in outP[p]) == 0
  [T2] UNMATCHED: u[i, k, w] <= scenarios[w]["s"][i][k] * y[P_index[S[i]]]
  [T3] UNMATCHED: quicksum(x[a, k, w] for a in inP[p]) - quicksum(x[a, k, w] for a in outP[p]) == 0
  [T4] UNMATCHED: quicksum(x[a, k, w] for a in outP[p]) <= M * y[P_index[p]]
  [T5] UNMATCHED: quicksum(x[a, k, w] for a in inC[C[j]]) + z[j, k, w] >= scenarios[w]["d"][j][k]
  [T6] UNMATCHED: quicksum(x[a, k, w] for a in inP[p]) <= M * y[P_index[p]]
GEN constraints (6):
  [G1] UNMATCHED: quicksum(x[j, i, k, w] for j in P if (j, i) in A) <= M * y[i]
  [G2] UNMATCHED: u[i, k, w] <= scenarios[w]['s'][i, k] * y[i]
  [G3] UNMATCHED: quicksum(x[i, j, k, w] for j in P if (i, j) in A) <= M * y[i]
  [G4] UNMATCHED: quicksum(x[i, j, k, w] for i, j in A if j == node) == quicksum(x[j, i, k, w] for i, j in A if i == node)
  [G5] UNMATCHED: quicksum(x[i, j, k, w] for i, j in A if j == node) + u[node, k, w] == quicksum(x[j, i, k, w] for i, j in A if i == node)
  [G6] UNMATCHED: quicksum(x[i, j, k, w] for i, j in A if j == node) + z[node, k, w] >= scenarios[w]['d'][node, k]
Matched objective:   0/1
Objective match (true -> gen): (none)
Unmatched TRUE objective(s):
  TRUE: c[i] * y[i] for i in range(len(P))) + quicksum( scenarios[w]["probability"] * ( quicksum( scenarios[w]["q"][a][k] * x[a, k, w] for a in range(len(A)) for k in range(len(K)) ) + quicksum( scenarios[w]["h"][j][k] * z[j, k, w] for j in range(len(C)) for k in r...
Unmatched GEN objective(s):
  GEN : c[i] * y[i] for i in P) + quicksum(scenarios[w]['probability'] * ( quicksum(scenarios[w]['q'][i, j, k] * x[i, j, k, w] for i, j, k in A for k in K if (i, j, k) in scenarios[w]['q']) + quicksum(scenarios[w]['h'][j, k] * z[j, k, w] for j in C for k in K if (j...
Partial score (eval1): 36.36%
Scores: P_total=36.36% P_extra=63.64% P_var=100.00% P_cons=0.00% P_obj=0.00%
Passed: 0/1
Accuracy: 0.00%
Partial score (eval1 fallback): 36.36
>> percent_extra_in_generated: 63.64%
>> percent_match_in_variables: 100.00%
>> percent_match_in_constraints: 0.00%
>> percent_match_in_objective: 0.00%
Compile error: 100.00%
Runtime error: 0.00%
\end{lstlisting}

\section{Conclusion}
\label{sec:conclusion}

Recent advances in large language models (LLMs) have created new opportunities to automate model formulation in operations research. While most prior studies have focused on deterministic (mixed-integer) linear programming models, this paper presents a systematic investigation of LLMs applied to Stochastic Optimization (SO), encompassing joint and individual chance-constrained programs, two-stage stochastic mixed integer linear programs (SMILP-2), and their deterministic mixed-integer linear programming (DMILP-2)  counterparts.

 A suite of customized prompting strategies, combining stepwise reasoning, chain-of-thought (CoT), and multi-agent (\texttt{agentic}) approaches, was designed to capture probabilistic feasibility in chance-constrained models and recourse logic in SMILP-2.
 
 A soft scoring metric was proposed to quantify algebraic and structural equivalence, addressing notational variation and permutation invariance that conventional metrics overlook. In addition, a standardized benchmark of natural language problem descriptions, input instances, and ground-truth Gurobi code was developed to ensure reproducibility and conduct experiments on SO problems.

 Extensive experiments across GPT-3.5, GPT-4, and GPT-4-Turbo variants revealed reasonable performance patterns and error behaviors across prompt designs, providing the first quantitative benchmark for LLM-based SO problem formulations.
Experimental results show that the full accuracy score was achieved by none of the GPT models or prompting methods. Nevertheless, the proposed scoring system paved the way to identify performance discrepancies among GPT models and prompting methods in this paper. GPT-4-Turbo outperformed all other models in most cases. It attained the highest partial scores for SMILP-2, DMILP-2, and joint chance-constrained problems, while GPT-3.5-Turbo-16k performed best on individual chance-constrained models.

Structured prompting improved formulation quality, with the \texttt{agentic} and \texttt{cot\_instructions} strategies achieving the best matching performance, whereas the standard baseline prompt frequently failed to produce complete or reliable results.

Future research will focus on extending this framework in several important directions:

 The whole proposed framework can be expanded to include a broader range of SO problems as well as contextual SO problems, distributionally robust and chance-recourse hybrids models, with a framework to customize structured prompts in accordance.
 
 In particular, we demonstrated that variants of prompts exhibit different performance depending on the problem type. There is a need to develop a tool that automatically analyzes a problem and renders the best prompts customized to that problem. Indeed, a prompt design problem requires rigorous probabilistic analysis based on the stochastic behavior of LLMs embedded inherently in them.    
 Hence, developing frameworks for designing efficient multi-agent structures can further improve the results. For instance, in our multi-agent framework, designing the architecture interactions and deciding the number of independent agents to maximize the performance is essential. Finally, the soft scoring can be enhanced to evaluate multi-stage stochastic programs and capture recursive decision structures. It can also be further developed to neutralize the effect of indexing errors in analyzing constraint and objective function correctness.

\section{Declaration of competing interest}

Authors affirm that there are no identifiable conflicting financial interests or personal associations that could be perceived as impacting the work described in this article.

\section{Data and code availability}

Data and code are available at \url{https://github.com/Amirreza-96/LLM-SO-manuscript}. 

\section{Acknowledgments}

The authors thank Dr. Güzin Bayraksan for her insightful and constructive feedback.

% The authors gratefully acknowledge the editor and anonymous reviewers for their valuable feedback and constructive suggestions. 
\bibliographystyle{plainurl}  
\bibliography{references}
\color{black}

\end{document}